\documentclass[journal]{IEEEtran}

\usepackage[utf8]{inputenc}
\usepackage{graphicx}
\usepackage[svgnames]{xcolor}
\usepackage{multirow}
\usepackage{hyperref}
\usepackage{amsmath} 
\usepackage{amssymb}  
\usepackage{lmodern}
\usepackage[most]{tcolorbox}
\usepackage{fontawesome5}
\newcommand{\ve}[1]{\ensuremath{\mathbf{#1}}}    

\newcommand{\ssl}[1]{\ensuremath{\mathcal{#1}}}  
\newcommand{\sr}[0]{{sim-to-real~}}
\usepackage{booktabs}
\usepackage{adjustbox}

\usepackage{color, colortbl}
\definecolor{Gray}{gray}{0.9}

\usepackage{pifont}

\DeclareUnicodeCharacter{0301}{\'{e}}

\usepackage[
natbib=true, 
backend=biber, 
sorting=none,
style=ieee,
date=year, 
giveninits=true,
url=false,
citestyle=numeric-comp,
maxbibnames=10]{biblatex}
\usepackage{color,soul}

\newtcolorbox{mybox}[5][]{%
        enhanced, width=7.9cm, height={#5},
        fontupper=\small\sffamily, fonttitle=\large\sffamily\bfseries\slshape,
        leftupper=1.1cm,
        leftrule=-1pt,
        colback=white,
        colframe=white,
        colupper=white,
        title=#2,
        attach title to upper={\\},
        underlay={\fill[#4] (frame.north west)--([xshift=-5mm]frame.north east)--(frame.east)--([xshift=-5mm]frame.south east)-|cycle;},
        overlay={\node[circle, minimum size=1.5cm, line width=1mm, draw=white, fill=#4, font=\Huge, text=white] at (frame.west) {#3};},
        #1, }

\hypersetup{
colorlinks = true,
linkcolor = {blue},
linkbordercolor = {white},
citecolor = {blue},
urlcolor = {blue}}
\usepackage{amsmath}
\begin{document}

\title{Survey of Learning-based Approaches for Robotic In-Hand Manipulation}

\author{Abraham Itzhak Weinberg$^{1}$, Alon Shirizly$^{2}$, Osher Azulay$^{3}$ and Avishai Sintov$^{3}$ 
\thanks{\noindent \\
$^{1}$ AI-WEINBERG AI Experts, Israel.\\
$^{2}$ Department of Mechanical Engineering, Technion - Israel Institute of Technology, Israel.\\
$^{3}$ School of Mechanical Engineering, Tel-Aviv University, Israel.}
}

\maketitle

\begin{abstract}
Human dexterity is an invaluable capability for precise manipulation of objects in complex tasks. The capability of robots to similarly grasp and perform in-hand manipulation of objects is critical for their use in the ever changing human environment, and for their ability to replace manpower. In recent decades, significant effort has been put in order to enable in-hand manipulation capabilities to robotic systems. Initial robotic manipulators followed carefully programmed paths, while later attempts provided a solution based on analytical modeling of motion and contact. However, these have failed to provide practical solutions due to inability to cope with complex environments and uncertainties. Therefore, the effort has shifted to learning-based approaches where data is collected from the real world or through a simulation, during repeated attempts to complete various tasks. The vast majority of learning approaches focused on learning data-based models that describe the system to some extent or Reinforcement Learning (RL). RL, in particular, has seen growing interest due to the remarkable ability to generate solutions to problems with minimal human guidance. In this survey paper, we track the developments of learning approaches for in-hand manipulations and, explore the challenges and opportunities. This survey is designed both as an introduction for novices in the field with a glossary of terms as well as a guide of novel advances for advanced practitioners. 
\end{abstract}

\begin{IEEEkeywords}
In-hand manipulation, Dexterous manipulation, Model learning, Reinforcement learning, Imitation learning
\end{IEEEkeywords}


\section{Introduction} 
\label{sec:introduction}

\begin{figure} 
\centering
\includegraphics[width=\linewidth]{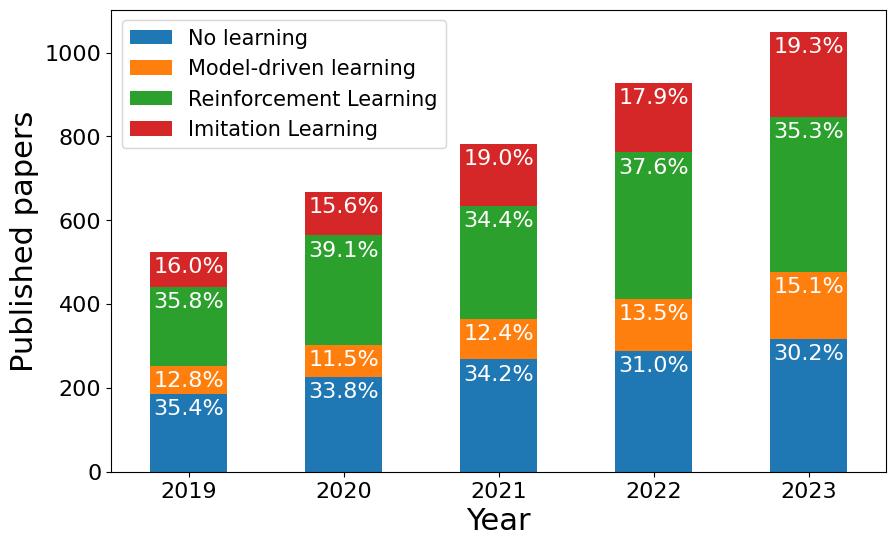}
\vspace{-0.6cm}
\caption{Statistics on paper publications which addressed or mentioned robotic in-hand manipulations over the past five years in three learning sub-fields: Model-driven learning, Reinforcement Learning (RL) and Imitation Learning (IL), along with papers that do not use any learning method. The search is based on Google Scholar and may include publications with merely a single mention of the topic and non-peer-review publications.}
\label{fig:stats}
\end{figure}

\begin{figure*}[!ht]
    \centering
		\includegraphics[width=\textwidth]{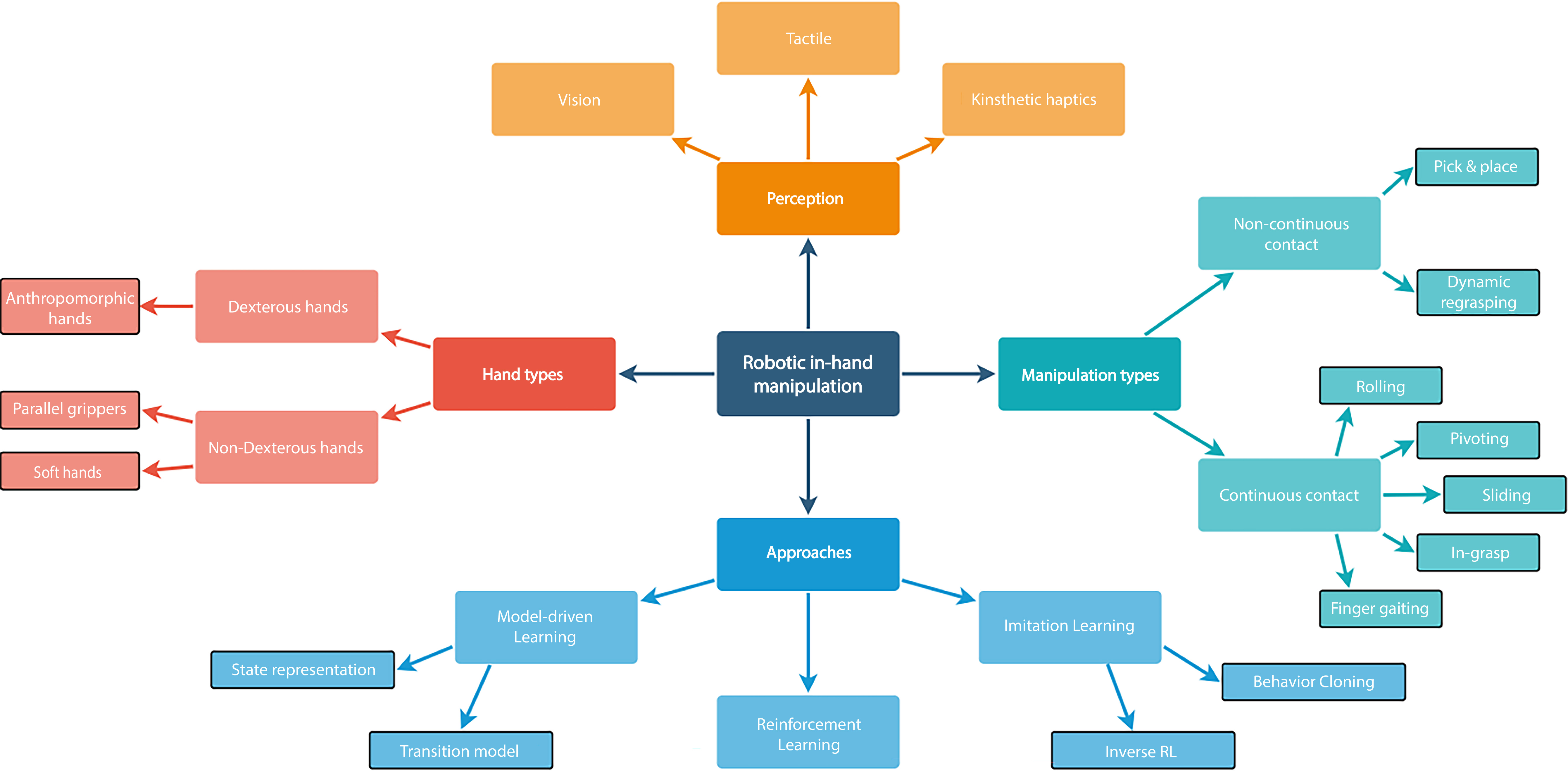}
	\caption{Taxonomy of robotic in-hand manipulation.}
	\label{fig:InHandTaxonmy}
\end{figure*}

Robot in-hand manipulation has long been considered challenging. However, it has undergone rapid development in recent years. With the vast industrial development and increase in demand for domestic usage, significant growth in interest in this field can be predicted. Evidently, we witness an increase in research papers, as shown in Figure \ref{fig:stats}, along with algorithms for solving versatile tasks. Just to mention a few, research has sought solutions for real-world tasks such as medical procedures \cite{Lehman2010}, assembly in production lines \cite{kang2021high}, and robotic assistance for sick and disabled users \cite{Petrich2021}. During the COVID-19 pandemic, there has been a significant need for autonomous and complex robot manipulators \cite{kroemer2021}. In this study, we survey various in-hand manipulation tasks of robotic hands and advanced learning approaches for achieving them. As commonly done by the robotics community and as shown in Figure \ref{fig:InHandTaxonmy}, we distinguish between two main categories of robot in-hand manipulation: \textit{dexterous} and \textit{non-dexterous} in-hand manipulations. To the best of our knowledge, we can conclude that the former approach is more prolific in algorithms and the number of published papers. In addition, we divide the types of manipulations into ones that have continuous and non-continuous contacts during execution. The continuous approach has more techniques than the other as it normally uses dexterous robotic hands with higher Degrees-Of-Freedom (DOF) in comparison to non-continuous approaches \cite{sun2021characterizing}.



Efforts for learning in-hand manipulation can be classified into three subfields: model-based methods, Reinforcement Learning (RL) and Imitation Learning (IL). Model-based methods focus on the supervised learning of the dynamics of a system or state representation. On the other hand, RL provides a reward function that embeds an implicit directive for the system to self-learn an optimal policy for completing a task. Similarly, IL requires a policy to imitate human expert demonstrations. These approaches have significant implications for in-hand manipulation, as they each offer unique advantages in improving robotic capabilities. They offer complementary contributions to improving robotic in-hand manipulation. Model-based methods provide a foundation for understanding system dynamics, RL enables self-learning of optimal policies, and IL allows for learning from human expertise. Figure \ref{fig:stats} shows the increase in paper publications over the past five years with regard to these three subfields. We note that the search is based on Google Scholar and aims only to show a trend. Results may include publications with merely a single mention of the topic without actual scientific contribution and non-peer-review publications. In this study, we survey in-hand manipulation approaches, tasks and applications that use either of these subfields with substantial contribution to the topic.


Previous surveys focused on specific aspects of robotic manipulation such as the use of contact \cite{Suomalainen2022}, space applications \cite{Papadopoulos2021}, handling of deformable objects \cite{Herguedas2019}, multi-robot systems \cite{Feng2020} and manipulation in cluttered environments \cite{Mohammed2022}. Some other surveys discussed learning approaches for general manipulation such as imitation learning \cite{Fang2019}, deep-learning \cite{Han2023} and general trends \cite{Billard2019,kroemer2021,Cui2021}. However, to the best of the author's knowledge, this is the first survey of learning approaches for robotic in-hand manipulation. Hence, this study offers multiple contributions. 
Papers were classified and grouped into meaningful clusters. The survey can help researchers to efficiently locate relevant research in a desired class and perceive what has already been achieved. Table \ref{tb:Pcomparison} provides a summary of prominent state-of-the-art work including key properties. These properties will be defined, introduced and discussed in the next section which provides an overview of in-hand manipulation. Practitioners can use our survey to estimate the added value of their research and compare it with previous studies. We also explain the relationships between the different subfields to showcase a wide perspective of the topic. In addition, our work can be seen as a survey of survey papers, similar to the references of some other survey papers. 

This study adopts a top-down approach. First, we provide a technical overview of in-hand manipulation, including the types of manipulation, hands and sensing modalities (Section \ref{sec:Overview}). This overview provides an understanding of the relevant hardware, manipulations and common terms to be used later. Next, we survey the subject from a high-level perspective and later zoom into more detailed sub-topics. In addition, we discuss popular tasks in each field. For each task, we often found several approaches while comparing the benefits of one over the other. Finally, we provide insights into future challenges and open problems that should be addressed by the robotics community. 

\begin{table*}[]
\centering
\caption{Summary of state-of-the-art work on learning approaches for in-hand manipulation (in alphabetical order)}
\label{tb:Pcomparison}
\begin{adjustbox}{width=\linewidth}
\begin{tabular}{l ccc c c c ccc}\toprule
\multirow{2}{*}{Paper} & \multicolumn{3}{c}{Learning type} & Manipulation & Dex./Non-Dex. & Rigid/Soft       & \multicolumn{3}{c}{Perception} \\\cline{2-4}\cline{8-10}
 & Model        & RL       & IL       &   type   & hand  & hand & Vision   & Tactile   & Kines.  \\\midrule

\rowcolor{Gray}
\citet{allshire2022transferring} & & \checkmark &  & In-Grasp & Dex. 
& Rigid & \checkmark & & \checkmark \\

\citet{antonova2017reinforcement} & & \checkmark & & Pivoting & Non-Dex. & Rigid & \checkmark & & \\

\rowcolor{Gray}
\citet{andrychowicz2020learning}\citet{OpenAI2019} & & \checkmark & & In-Grasp & Dex. & Rigid & \checkmark & & \checkmark \\

\citet{Arunachalam2023,Arunachalam2023holo} & & & \checkmark & Mix & Dex. & Rigid & \checkmark & & \\

\rowcolor{Gray}
\citet{azulay2022learning} & \checkmark & & & In-grasp & Non-Dex. & Soft & & \checkmark  & \checkmark\\

\citet{azulay2022haptic} & & \checkmark & & In-Grasp & Non-Dex. & Soft & & & \checkmark\\

\rowcolor{Gray}
\citet{Calli2018} & \checkmark & & & In-Grasp & Non-Dex & Soft & \checkmark & & \checkmark \\

\citet{chen2021simple,chen2022} & & \checkmark & & Mix & Dex. & Rigid & \checkmark & & \\

\rowcolor{Gray}
\citet{Chen2023} & & \checkmark & & F. Gait & Dex. & Rigid & \checkmark & & \checkmark \\

\citet{Cruciani2017,Cruciani2018} & & \checkmark & & Pivoting & Non-Dex. & Rigid & \checkmark & & \\

\rowcolor{Gray}
\citet{deng2020learning} & & & \checkmark & Mix & Dex. & Rigid & \checkmark & & \checkmark \\

\citet{Dimou2023} & \checkmark & & & In-Grasp & Dex. & Rigid & \checkmark & & \\

\rowcolor{Gray}
\citet{Falco2018} & & \checkmark & & In-Grasp & Dex. & Soft & \checkmark & \checkmark &  \\

\citet{funabashi2019morphology,funabashi2020variable,Funabashi2019} & & \checkmark & & Rolling & Dex. & Rigid 
& & \checkmark & \checkmark 
\\

\rowcolor{Gray}
\citet{GarciaHernando2020} & & & \checkmark & Mix & Dex & Rigid & \checkmark & & \\

\citet{Gupta2021} & & \checkmark & &  Mix & Dex. & Rigid & \checkmark &  &  \\ 

\rowcolor{Gray}
\citet{Handa2023} & & \checkmark & & Mix & Dex. & Rigid & \checkmark & & \\ 

\citet{Huang2021} & & \checkmark & & F. Gait & Dex. & Rigid & \checkmark &  & \\

\rowcolor{Gray}
\citet{Jain2019} & & \checkmark & \checkmark & In-Grasp & Dex. & Rigid & \checkmark & \checkmark & \checkmark \\

\citet{Khandate2022,Khandate2023} & & \checkmark & & F. Gait & Dex. & Rigid & & \checkmark & \checkmark \\

\rowcolor{Gray}
\citet{kimmel2019belief} & \checkmark & & & In-Grasp & Non-Dex. & Soft & \checkmark & & \checkmark \\

\citet{korthals2019multisensory} & & \checkmark & & Mix & Dex. & Rigid & \checkmark & \checkmark & \checkmark \\

\rowcolor{Gray}
\citet{Kumar2016b} & & & \checkmark & Mix & Dex. & Rigid & & \checkmark & \checkmark \\

\citet{Li2014} & & & \checkmark & In-grasp & Dex. & Rigid & \checkmark & \checkmark & \checkmark \\

\rowcolor{Gray}
\citet{Li2020} & \checkmark & \checkmark & & In-Grasp & Dex. & Rigid & \checkmark & & \checkmark 
\\ 

\citet{luo2023} & \checkmark & & & Mix & Dex. & Rigid & \checkmark & & \checkmark \\

\rowcolor{Gray}
\citet{melnik2021} & & \checkmark & & Mix & Dex. & Rigid & \checkmark & \checkmark & \checkmark \\

\citet{morgan2020object} & \checkmark & & & In-Grasp & Non-Dex. & Soft & \checkmark & &  \\

\rowcolor{Gray}
\citet{Morgan2021} & \checkmark & \checkmark & & Mix & Non-Dex. & Soft & \checkmark & &  \\

\citet{nagabandi2020deep} & \checkmark & \checkmark & & Mix & Dex. & Rigid & \checkmark & & \\

\rowcolor{Gray}
\citet{Orbik2021} & & & \checkmark & Mix & Dex. & Rigid & \checkmark & & \checkmark \\

\citet{Park2024} & \checkmark & & & In-Grasp & Non-Dex. & Soft & & \checkmark & \\

\rowcolor{Gray}
\citet{Pitz2023} & & \checkmark & & In-Grasp & Dex. & Rigid & & & \checkmark \\

\citet{qi2022} & & \checkmark & & F. Gait & Dex. & Rigid & & & \checkmark \\

\rowcolor{Gray}
\citet{Qi2023} & & \checkmark & & F. Gait & Dex. & Rigid & \checkmark & \checkmark & \checkmark \\

\citet{radosavovic2020state} & & & \checkmark & Mix & Dex. & Rigid & \checkmark & & \\

\rowcolor{Gray}
\citet{Rajeswaran2017} & & & \checkmark & Mix & Dex. & Rigid & \checkmark & & \checkmark \\

\citet{Shin2024} & & & \checkmark & Rolling & Dex. & Rigid & \checkmark & \checkmark & \checkmark \\

\rowcolor{Gray}
\citet{Sievers2022} & & \checkmark & & In-Grasp & Dex. & Rigid & & & \checkmark \\

\citet{Sintov2019,sintov2020motion} & \checkmark & & & In-Grasp & Non-Dex. & Soft & \checkmark & & \checkmark \\

\rowcolor{Gray}
\citet{Solak2019} & & & \checkmark & In-Grasp & Dex. & Rigid & & & \checkmark \\

\citet{Solak2023} & & & \checkmark & In-Grasp & Dex. & Rigid & & \checkmark & \checkmark \\

\rowcolor{Gray}
\citet{srinivasan2020learning} & & \checkmark & & In-Grasp & Dex. & Rigid & \checkmark & & \checkmark \\ 

\citet{Tao2023} & & \checkmark & & Mix & Dex. & Rigid & \checkmark & & \\

\rowcolor{Gray}
\citet{Toskov2023} & \checkmark & & & Pivoting & Non-Dex. & Rigid & & \checkmark & \\

\citet{Toledo2021} & & \checkmark & & Pivoting & Non-Dex. & Rigid & \checkmark & &  \\

\rowcolor{Gray}
\citet{VanHoof2015} & & \checkmark & & In-Grasp & Non-Dex. & Soft & & \checkmark & \\

\citet{Veiga2020} & & \checkmark & & In-Grasp & Dex. & Rigid & & \checkmark & \\

\rowcolor{Gray}
\citet{wang2020swingbot} & \checkmark & & & Pivoting & Non-Dex. & Rigid &  &  & \checkmark \\

\citet{Wei2023} & & & \checkmark & Sliding & Dex. & Rigid & \checkmark & & \\

\rowcolor{Gray}
\citet{Yang2023} & & \checkmark & & Mix & Dex. & Rigid & & \checkmark & \\

\citet{Yin2023} & & \checkmark & & Mix & Dex. & Rigid & & \checkmark & \checkmark \\

\rowcolor{Gray}
\citet{Yuan2020} & & & \checkmark & Rolling & Non-Dex. & Rigid & \checkmark & & \checkmark \\

\citet{Yuan2023} & & \checkmark & & Mix & Dex & Rigid & \checkmark & \checkmark & \checkmark \\

\bottomrule                  

\end{tabular}
\end{adjustbox}

\end{table*}



\section{Overview on in-hand manipulations} 
\label{sec:Overview}

Robotic in-hand manipulation involves physical interaction between a robotic end-effector, an object and often with the environment \cite{Cruciani2020}. The properties of an end-effector define its ability to manipulate the object including: sensory perception, number of DOF, kinematics and friction. In this section, we provide an overview of various types of in-hand manipulations and the robotic hand types that are capable to exert them. In addition, we discuss the common perception and control methods used in these manipulations.


\subsection{Dexterous and non-dexterous manipulation}

The conventional paradigm is to distinguish between dexterous and non-dexterous hands. Generally, dexterous manipulation is the cooperation of multiple robot arms or fingers to manipulate an object \cite{M.Okamura2000}. Dexterous in-hand manipulation is, therefore, the manipulation of an object in the hand by using its own mechanics \cite{Mason1985}. Naturally, dexterous in-hand manipulation requires a high number of DOF and includes, in most cases, anthropomorphic hands. Contrary to dexterous hands, non-dexterous ones have a low number of DOF and, thus, do not have an intrinsic capability to manipulate objects by themselves and require some extrinsic involvement. 

\begin{figure}
    \centering
    \begin{tabular}{cc}
    \includegraphics[width=0.48\linewidth]{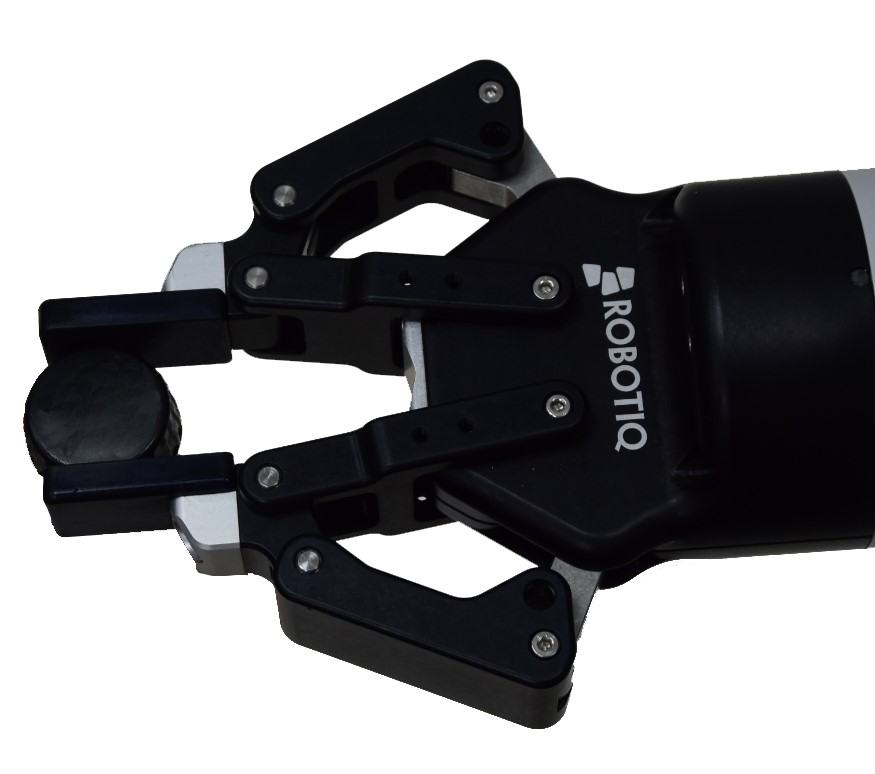} & \includegraphics[width=0.42\linewidth]{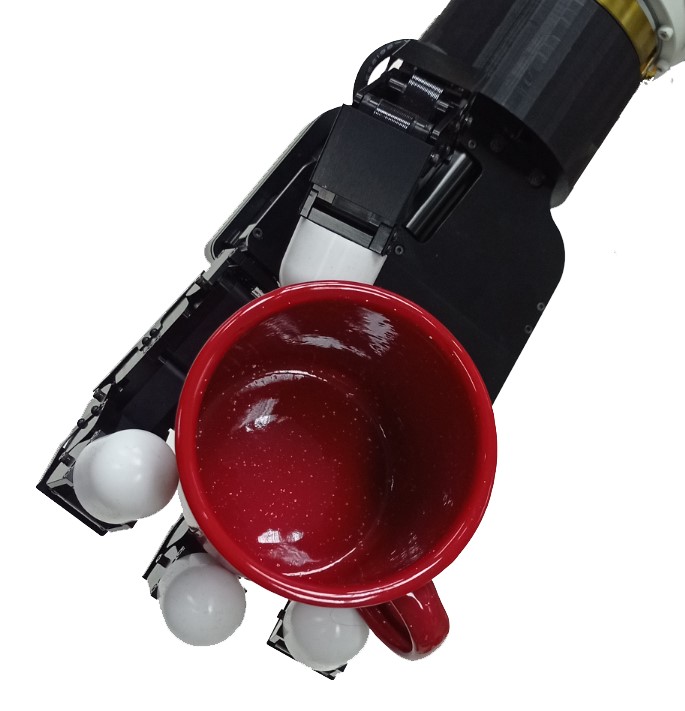} \\
    (a) & (b) \\
    \includegraphics[width=0.32\linewidth]{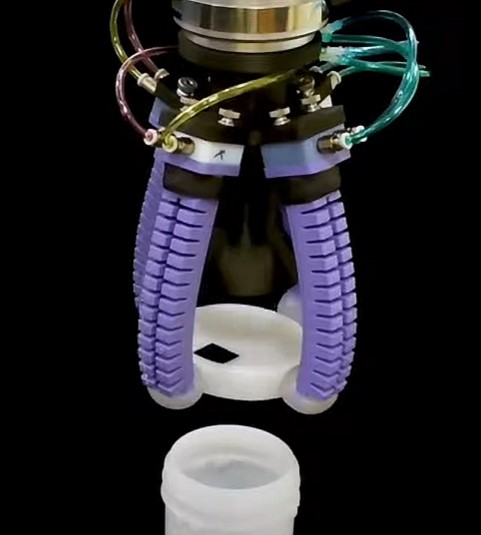} & \includegraphics[width=0.48\linewidth]{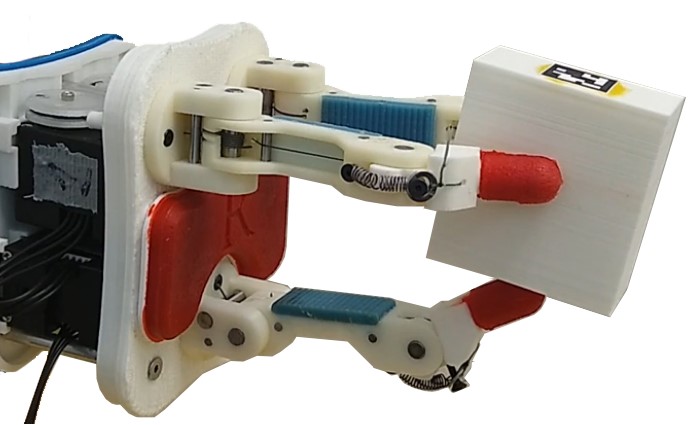} \\
    (c) & (d) \\
    \end{tabular}
    \caption{Various dexterous and non-dexterous hands. (a) Non-dexterous parallel jaw gripper model 2F-85 by Robotiq. The gripper has only a single DOF for opening and closing on an object. (b) The four-finger dexterous and anthropomorphic Allegro hand with 16 DOF. (c) A four finger non-dexterous soft hand operated by pneumatic bending actuators \cite{Abondance2020}. (d) Underactuated compliant hand model-O from the Yale OpenHand project \cite{Ma2017YaleOP}. Images (a),(b) and (d) were taken by the authors.}
    \label{fig:hands}
\end{figure}


\subsection{Types of in-hand manipulations}

We now introduce the types of in-hand manipulations commonly addressed in the literature and distinguish between those that maintain and do not maintain continuous contact with the object. These types can be referred to as both dexterous and non-dexterous manipulations as will be discussed later.

\subsubsection{In-hand manipulations that maintain continuous contact}
Initiating object motion within a robotic hand poses some risk of losing control and potentially dropping it. Hence, the majority of in-hand manipulations perform the motion while maintaining sufficient contact with the object, as it is the safest approach. Nevertheless, a prominent condition for a successful manipulation is that grasp stability is guaranteed throughout the motion. Following are the key in-hand manipulation types that maintain contact.

\begin{itemize}
    
    \item \textit{Rolling.} Rolling manipulation is the ability to rotate an object within the hand through rolling contact. This is caused by finger motion during contact while the object is held against another part of the hand, usually static, such as another finger or palm \cite{Han1997}. Rolling manipulation is limited to objects of certain geometries and is most feasible with round ones. 
    
    \item \textit{Pivoting.} Pivoting is the reorientation of an object between two fingers with respect to the hand \cite{Sintov2016,bhatt2022surprisingly}. The rotation point is commonly the pinching point of the fingers and the reorientation is conducted to some desired angle. The pivoting operation can be done by utilizing gravity \cite{Vina2015}, initiating external contact \cite{Dafle2014} or generating dynamic motions of the robotic arm \cite{SintovSwingUp2016, Cruciani2017}.
    
    \item \textit{Sliding.} In in-hand sliding manipulation, controlled slip is initiated in order to vary the relative position of the object with regards to the hand \cite{shi2017dynamic}. The object slides along the links of the hand to a desired position due to forces from the fingers or external forces \cite{Shi2020}. 
    
    \item \textit{In-Grasp.} While pivoting, sliding and rolling aim to change either position or orientation of the object, in-grasp exploits kinematic redundancy of the hand to vary both the position and orientation of the object \cite{Ma2011}. During manipulation, finger contacts and a stable grasp are maintained. However, sliding or contact rolling may occur \cite{Calli2018}. 
    
    \item \textit{Finger Gaiting.} An approach analogous to gait where the hand’s DOF are exploited to switch between contact locations while maintaining a force-closure grasp \cite{Rus1999,Xu2007,Fan2017,bhatt2022surprisingly}. Hence, the approach is commonly performed quasi-statically where the motion is performed relatively slow to reduce dynamic effects. Finger gaiting may be considered quite wasteful, as it requires sufficiently many DOFs to manipulate the grasped object between two grasp configurations while maintaining stable grasps.  
\end{itemize}

\subsubsection{In-hand manipulations that do not maintain continuous contact}
\begin{itemize}
    
    \item \textit{Pick and place.} While not usually considered as in-hand manipulation, pick-and-place is worth mentioning since it is the most common. The method designates a work area near the robotic arm, where the grasped object can be placed in a controlled manner and then picked up again at a new grasp configuration \cite{Lozano-Perez1987,Tournassoud1987}. This approach consumes valuable production time and occupies a substantial work area.
    
    \item \textit{Dynamic Regrasping.} In this approach, the robot initiates an intended loss of grasp stability through the set of dynamical motions. In most cases,   
    the object is thrown or released into midair and later caught in different grasp configurations. Hence, the hand loses contact (fully or partly) with the object and regains contact by catching it at the final contact points \cite{SintovInHand2017}. Such a method has the advantage of fast manipulation and may require a low number of DOF. However, in contrast to manipulations that maintain contact, the success rate for dynamic regrasping may be lower as object stability is not maintained throughout the motion.
    
\end{itemize}


\subsection{In-hand manipulation with Non-dexterous hands}

\subsubsection{Parallel grippers} 

The most common and ubiquitous non-dexterous robotic hand is the \textit{parallel or jaw gripper} seen in Figure \ref{fig:hands}a. Parallel grippers are widely used due to their simplicity, durability and low cost. They can precisely grasp almost any object of the same scale and, therefore, are ubiquitous in industrial applications of material handling \cite{Guo2017}. Parallel grippers normally have only one DOF for opening and closing the jaws. Hence, they do not have independent in-hand manipulation capabilities. Consequently, solutions for in-hand manipulation with parallel grippers often involve the discrete manipulation approach of \textit{pick-and-place} \cite{Tournassoud1987}. In pick-and-place, the object is placed on a surface and picked up again in a different grasp configuration \cite{Zeng2018}. However, the picking and placing can be slow and demands a large surface area around the robot. Hence, approaches for in-hand manipulation with parallel grippers, that do not involve picking and placing, are also divided to \textit{extrinsic and intrinsic dexterity} \cite{Ma2011,Billard2019}. The former compensates for the lack of gripper DOF and involves actions of the entire robotic arm for either pushing the object against an obstacle \cite{Dafle2014,Chavan-Dafle2020} or performing dynamic manipulation. For instance, pivoting can be done by intrinsic slippage control or extrinsic dynamic manipulation of the arm \cite{Vina2015, SintovSwingUp2016, Cruciani2018}. Slippage control leverages gravity and tunes finger contact force of the parallel gripper \cite{Costanzo2021}. \citet{Costanzo2021b} exploited a dual-arm system and tactile feedback to allow controlled slippage between the object and parallel grippers. The work of \citet{Shi2017} controlled the force distribution of a pinch grasp to predict sliding directions. Similarly, \citet{Chen2021} controlled the sliding velocity of an object grasped by a parallel gripper. 

In intrinsic manipulation, the available DOF of the gripper are exploited for manipulating the grasped object \cite{Cruciani2018b}. While jaw grippers have only one DOF, some work has been done to augment their intrinsic manipulation capabilities. These robotic hands, equipped with additional functionalities beyond the traditional single DOF parallel gripper, can no longer be categorized as simple grippers. Seminal work by \citet{Nagata1994} proposed six gripper mechanisms with an additional one DOF at the tip, each having the ability to either rotate or slide an object in some direction. Similarly, a passively rotating mechanism was integrated into the fingers of the gripper allowing the object to rotate between the fingers by gravity \cite{Terasaki1998}. \citet{Zhao2020} augmented a jaw gripper tip with a two DOF transmission mechanism to re-orient and translate randomly placed screws. \citet{Zuo2021} added a linear actuation along each of the two fingers to enable translation and twist of a grasped object. Similarly, a rolling mechanism was added to the gripper by \citet{Chapman2021} in order to manipulate a flat cable. In-hand manipulation was also enabled for a minimal underactuated gripper by employing an active conveyor surface on one finger \cite{Ma2016}. \citet{Taylor2020} included a pneumatic braking mechanism in a parallel gripper in order to transition between object free-rotation and fixed phases. The above augmentation methods for parallel grippers are limited to one manipulation direction and yield bulky mechanisms that complicate the hardware. However, a simple vibration mechanism was recently proposed to enable $SE(2)$ sliding motion of a thin object between the jaws of a parallel gripper \cite{Nahum2022}. These innovative designs offer enhanced in-hand manipulation capabilities without requiring complex additional controls or software, opening up new avenues for research and development within the field of in-hand manipulation. Traditional control and motion planning methods for these systems often lack the flexibility to generalize to diverse tasks and objects. Learning-based approaches offer a promising solution for enhancing the capabilities of these mechanisms.


\subsubsection{Soft hands}

Soft hands are robotic grippers that are comprised of soft or elastic materials. Due to their soft structure, they usually provide passive compliance upon interaction with the environment \cite{Zhou2018}. Hence, they can grasp objects of varying sizes and shapes without prior knowledge. A class of soft hands is the \textit{pneumatic-based hands} where stretchable fingers can be inflated to generate a grasp. For instance, RBO Hand 2 is a compliant, under-actuated anthropomorphic robotic hand \cite{Deimel2014}. Each finger of the hand is made of cast silicon wrapped with inelastic fabric. When inflated, the fabric directs the stretch of the fingers toward a compliant grasp. 
In-hand manipulation with pneumatic-based hands was demonstrated for which heuristic finger gait enabled continuous object rotation (Figure \ref{fig:hands}c) \cite{Abondance2020}. Another pneumatic hand with reconfigurable fingers and an active palm was designed to enable in-hand dexterity while maintaining low mechanical complexity \cite{Pagoli2021}. \citet{Batsuren2019} presented a soft robotic gripper for grasping various objects by mimicking in-hand manipulation. It consists of three fingers, where each of them contains three air chambers: two side chambers for twisting in two different directions and one middle chamber for grasping. The combination of these air chambers makes it possible to grasp an object and rotate it.

An important class commonly referred as a soft hand is the \textit{Underactuated} or \textit{Compliant} hand \cite{Dollar2010,Liu2020}. While the links of such a hand are generally rigid, each finger has compliant joints with springs where a tendon wire runs along its length and is connected to an actuator (Figure \ref{fig:hands}d). Such a structure enables a two or more finger hand to passively adapt to objects of uncertain size and shape through the use of compliance \cite{Odhner2011}. They can, therefore, provide a stable and robust grasp without tactile sensing or prior planning, and with open-loop control. In addition, due to the low number of actuators, they enable a low-cost and compact design. Recently, open-source hardware was distributed for scientific contributions and can be easily modified and fabricated by 3D printing \cite{Ma2017YaleOP}. Along with good grasping capabilities, precise in-grasp manipulation was shown possible \cite{odhner2015stable}. Using visual servoing along with a linear approximation of the hand kinematics, closed-loop control of a two-finger hand was demonstrated \cite{calli2016vision} and later used to track paths planned with an optimization-based model-free planner \cite{calli2018path}. However, a precise analytical model for soft hands is not easy to acquire due to the compliance and inherent fabrication uncertainties. Therefore, data-based models were proposed and will be discussed later.


\subsection{In-hand manipulation with Dexterous hands}

\citet{Mason1985} claimed that rigid hands can acquire controllability of an object with at least three fingers of three joints each. Such a hand control is termed dexterous manipulation and the hand is a \textit{dexterous hand}. Naturally, grippers that satisfy this dexterity condition are bio-inspired or anthropomorphic \cite{LlopHarillo2019} (Figure \ref{fig:hands}b). Early work on dexterous anthropomorphic hands includes a three-finger and 11-DOF hand \cite{okada1979object}, the four-finger Utah/MIT hand \cite{Jacobsen1986}, and later the Barrett and DLR hands \cite{Townsend2000,Butterfass2001}. Furthermore, extensive work was done on five-finger anthropomorphic hands. Similar to the DLR hand, The Gifu Hand used 16 built-in servo-motors in the joints \cite{Mouri2002}. On the other hand, the Robotnaut hand was designed for space usage and included flex shafts for bending the fingers \cite{Lovchik1999}. A hand from Karlsruhe used 13 flexible fluidic actuators for a lightweight design \cite{Schulz2001}. The UB hand is a five-finger anthropomorphic hand that used elastic hinges to mimic human motion \cite{Lotti2004}. Beyond anthropomorphic designs, few non-anthropomorphic dexterous hands have been proposed, incorporating multiple fingers in various designs \cite{Hammond2012}. However, most attempts to design a non-anthropomorphic multi-finger hand adhere to under-actuation, limiting their dexterity \cite{Molnar2022}.

While recent work on in-hand manipulation with dexterous hands is based on learning approaches, earlier and few recent ones have proposed non-data-driven methods. For instance, the work by \citet{Furukawa2006} proposed a high-speed dynamic regrasping strategy with a multi-fingered hand based on visual feedback of the manipulated object. A different work introduced a planning framework for an anthropomorphic hand to alternate between finger gait and in-grasp manipulations \cite{Sundaralingam2018}. Recent work by \citet{Pfanne2020} used impedance control for stable finger gaiting over various objects with a dexterous multi-finger hand.

Multi-finger anthropomorphic hands are commonly employed in the development of bionic prostheses as they resemble the human hand \cite{Cordella2016}. They are usually operated by Electromyographic (EMG) signals to reduce the cognitive burden on the user \cite{Starke2022}. While these hands are often highly dexterous and have multimodal information from various sensors \cite{Stefanelli2023}, their use is commonly limited to pick and place tasks \cite{Marinelli2022}. Hence, the learning methods explored in this paper offer potential avenues for advancing the capabilities of various hands including prosthetic ones with in-hand manipulation tasks.


\subsection{Perception}

Humans use both visual feedback and touch perception for interacting with the environment and, in particular, manipulate objects within their hand \cite{RoblesDeLaTorre2001}. Such sensory modules have been widely explored in robotics, both individually and combined. 

\subsubsection{Vision}

Different variations of visual perception are used to observe a manipulated object and estimate its pose in real time. The easiest application is the positioning of \textit{fiducial markers} such as ArUcO \cite{GarridoJurado2014}, AprilTags \cite{Olson2011} or reflective markers for a Motion Capture system (MoCap) \citet{azulay2022learning}. These markers provide instant pose recognition of a rigid object without the need for its geometry recognition \cite{Kalaitzakis2021}. However, the requirement to apply them on an object prevents spontaneous unplanned interaction with an object. Specifically with reflective markers, the work is limited to a room or lab where the MoCap system is located. In general, vision-based markers are required to be continuously visible to the camera. Hence, they are commonly for manipulation of specific known objects, or for prototyping. For instance, fiducial markers were used in visual servoing \cite{Calli2017} and hand state representation \cite{Sintov2019} during in-grasp manipulation of an object with an underactuated hand.

While fiducial markers offer immediate pose estimation, their reliance on predefined visual patterns limits their applicability in real-world environments. To address this, visual perception, combined with learning-based methods, is often employed for robust object recognition and pose estimation.
Visual pose estimation, which is based on geometry recognition of the object, is usually based on an RGB (monocular) camera, depth camera or both (RGB-D). With RGB data, much work has been done to regress 2D images to spatial pose of objects \cite{Rad2017,Billings2019,Kokic2019}. Nevertheless, in simpler applications where the object is known, it can be segmented using image processing tools. For instance, a high-speed vision system was used by \citet{Furukawa2006} to track a cylinder thrown and caught by a multi-finger hand. Similarly, a high-speed camera was used to solve a Rubik's cube with a fast multi-finger hand \cite{Higo2018}. A work by OpenAI used three RGB cameras to train a model for pose estimation of a cube manipulated by the Shadow hand \cite{andrychowicz2020learning}. \citet{ichnowski2021dex} presented Dex-NeRF, a novel approach that enables grasping using Neural Radiance Field (NeRF) technique. NeRF receives five-dimensional vectors as input and can be used for grasping transparent objects. The RGB values are calculated using an Artificial Neural Network (ANN) only after the initial stages.

Contrary to RGB cameras, 3D sensing such as stereo cameras, laser scanners and depth cameras enable direct access to the distance of objects in the environment. RGB-D sensing, in particular, provides an additional point cloud corresponding to the spatial position of objects in view. Commonly used depth cameras include Intel's RealSense and StereoLab's ZED, where the latter  leverages GPU capabilities for advanced spatial perception. For instance, an RGB-D camera was used to estimate the pose of objects before and during grasp by a soft hand \cite{Choi2017}. Similar work involved a depth camera to demonstrate robust pose estimation of objects grasped and partly occluded by a two-finger underactuated hand \cite{Wen2020}. Although visual perception can provide an accurate pose estimation of a manipulated object, it requires a line of sight. Hence, it cannot function in fully occluded scenes and may be sensitive to partial occlusions. Haptic-based approaches can, therefore, provide an alternative or complementary solution. 

\subsubsection{Haptics}

Information from haptic sensors is acquired through direct contact with objects by either tactile sensing \cite{yousef2011tactile} or internal sensing of joint actuators known as Kinesthetic (or Proprioception) haptics \cite{Carter2005}. Traditionally, tactile refers to information received from touch sensing, while kinesthetic refers to internal information of the hand sensed through movement, force or position of joints and actuators. While kinesthetic haptics can be easier to measure, tactile sensing is the leading haptic-based sensing tool for object recognition and in-hand manipulation. State-of-the-art tactile sensors include force sensors on fingertips, arrays of pressure sensors \cite{Bimbo2016} or high-resolution optical sensors \cite{yuan2017gelsight,Sun2022}. With these sensors, robotic hands can continuously acquire information about the magnitude and direction of contact forces between them and the manipulated object during interaction. An array of pressure sensors was used for servo control of the Shadow hand in in-hand manipulation tasks of deformable objects \cite{Delgado2017}. Optical tactile sensors work by projecting a pattern of light onto a surface and observing the distortion of that pattern caused by contact using an internal camera. Different sensors utilize different cameras with sensing resolution of up to $2592\times1944$. This distortion provides information about the shape, texture, and pressure applied to the surface \cite{taylor2022gelslim,Azulay2024}. \citet{lambeta2020digit} used these sensors to learn in-hand manipulation models. To combine the advantages of haptics and visual perception, some work has been done with both to explore the hand-object interaction during in-grasp manipulation \cite{He2015}. While haptics provides valuable information regarding the state of contact with the environment, traditional analytical methods are often insufficient for processing this data. Hence, learning-based approaches have emerged as a promising solution for extracting meaningful information from haptic sensor data.

\subsection{Simulation of In-hand Manipulation}

Simulating in-hand and dexterous manipulation is a critical aspect of robotic research, offering a controlled environment for developing and testing advanced control algorithms. High-fidelity simulators like MuJoCo \cite{todorov2012mujoco} and Isaac Gym \cite{makoviychuk2021isaac} allow researchers to model complex interactions between robotic hands and objects, enabling the study of tasks such as reorienting a cube \cite{Andrychowicz2017}, opening doors \cite{Rajeswaran2017} or dynamically adjusting grasps on irregular objects \cite{agarwal2023dexterous}. For instance, MuJoCo's ability to model soft contacts and Isaac Gym's high-speed parallel simulations make them valuable tools for training and evaluating robotic manipulation strategies. The use of simulations in dexterous manipulation research is invaluable. It enables large-scale experimentation and rapid iteration, eliminating the risks or costs associated with physical testing. Researchers can explore complex manipulation tasks with multimodal sensing, including tactile and visual inputs \cite{Yuan2023}, in a controlled and scalable setting. Ultimately, these simulations drive the development of more adaptive robotic systems capable of human-like dexterity in unstructured environments.

Despite their advantages, these simulators face significant challenges in accurately replicating real-world physics, particularly in modeling friction, soft deformation and contact forces \cite{haldar2023teach}. This sim-to-real gap can result in behaviors learned in simulation failing to transfer seamlessly to physical robots due to unmodeled dynamics and sensor noise. Furthermore, while rigid body dynamics are often well-represented, simulators struggle with soft materials and deformable objects, which is crucial for tasks like manipulating cloth or delicate items. More specifically, the simulation of underactuated hands is still a challenge.

\subsection{Datasets of in-hand manipulation motions}

Learning models require a significant amount of data to achieve sufficient accuracy. Data in many applications is inherently high-dimensional, often consisting of multimodal signals like visual and haptic data. Simulators, such as mentioned above, provide an environment to collect such data. However, the reality gap is often too large and the acquisition of real-world data is necessary. However, acquiring the data may be exhausting, expensive and even dangerous. Hence, practitioners often disseminate their collected data for the benefit of the community and for potential benchmarking \cite{khazatsky2024}. For example, RealDex is a dataset focused on capturing authentic dexterous hand motions with human behavioral patterns based on tele-operation \cite{Liu2024RealDexTH}. The RUM dataset includes data of real in-hand manipulation of various objects with adaptive hands \cite{Sintov2020a}. A prominent dataset is the YCB object and model set \cite{Calli2015}, aimed to provide a standard set of object for benchmarking general manipulation tasks including in-hand ones \cite{Cruciani2020}. Some datasets are simulation based such as the DexHand \cite{Nematollahi2022} where the data is comprised of RGB-D images of a Shadow Hand robot manipulating a cube. Overall, publicly available datasets are an important tool to promote standardized objects, tasks and evaluation metrics to benchmark and compare different approaches to robotic in-hand manipulation.

\section{Model-Driven Learning for In-hand Manipulation}

The establishment of control policies for in-hand manipulation remains challenging regardless of gripper, object or task properties. Various contact models and hand configurations have been used in the literature to develop kinematic and dynamic models for in-hand manipulation as described in previous sections. In order to execute in-hand manipulation tasks with these models, detailed knowledge of the object-hand interaction is required. For most robotics scenarios, however, such information cannot be reasonably estimated using conventional analytical methods since precise object properties are often not a priori known. 
Model learning offers an alternative to careful analytical modeling and accurate measurements for this type of system, either through robot interaction with the environment or human demonstrations. Learning a model can be done explicitly using various Supervised Learning (SL) techniques, or implicitly by maximizing an objective function. In this section, we will focus on the former technique of supervised learning while the latter will be covered in the following section.


\subsection{Learning state representation} 

Learning a state representation for in-hand robotic manipulations refers to the process of developing a mathematical model that describes the various states of the hand-object system during manipulation. This model can be used to represent the position, orientation, velocity, and other physical properties of an object. Furthermore, the model can be used to predict object response to certain actions. 

State representation is an important building block where the object-hand configuration is sufficiently described at any given time. For example, if a robot is trying to roll an object within the hand, it may use some state representation to measure and track the object's pose, and use this information to exert informed actions. In SL, an ANN is commonly used to extract relevant features from data and learn useful features from high-dimensional observation spaces \cite{azulay2022learning, andrychowicz2020learning, Funabashi2019,Sodhi2020,Dimou2023}. It is also effective in combining data from multiple sensors or information sources \cite{qi2022}, and is often used by robots to merge information from different modalities, such as vision and haptic feedback \cite{andrychowicz2020learning}. Without a compact and meaningful representation of the object-hand state, the robot may struggle to perform successful and efficient manipulations \cite{azulay2022learning}.

Haptic perception is commonly used to learn various features of an object in uncertain environments so as to grasp and manipulate it. Such information may include stiffness, texture, temperature variations and surface modeling \cite{Su2012}. Often, haptic perception is used alongside vision to refine initial pose estimation \cite{Bimbo2013}. Contact sensing is the common approach for pose estimation during manipulation \cite{azulay2022learning}. Such sensing has been generally achieved using simple force or pressure sensors \cite{Tegin2005, cheng2009novel, wettels2009multi}. As such, \citet{Koval2013} used contact sensors and particle filtering to estimate the pose of an object during contact manipulation. \citet{Park2024} used soft sensors in a pneumatic finger and a neural-network to estimate the angles of the finger. In recent years, optical sensor arrays have become more common due to advancements in fabrication abilities and due to their effectiveness in covering large contact areas \cite{Bimbo2016}. The softness of the sensing surface allows the detection of contact regions as it deforms while complying with the surface of the object. Changes in images captured by an internal camera during contact are analyzed. 

Several works have used data from optical-based tactile sensors and advanced deep-learning networks to estimate the relative pose of an object during contact manipulation. For example, \citet{Sodhi2020} used data from an optical-based tactile sensor to estimate the pose of an object being pushed, while others \cite{Lepora2020, psomopoulou2021robust} explored the use of these sensors for estimating the relative pose of an object during a grasp. In a work by \citet{wang2020swingbot}, the use of tactile sensing for in-hand physical feature exploration was also explored in order to achieve accurate dynamic pivoting manipulations. \citet{wang2020swingbot} also used optical tactile sensors on a parallel gripper in order to train a model to predict future pivoting angles given some control parameters. \citet{Toskov2023} addressed the pivoting with tactile sensing and trained a recursive ANN for estimating the state of the swinging object. The model was then integrated with the gripper controller in order to regulate the gripper-object angle. \citet{Funabashi2019} learned robot hand-grasping postures for objects with tactile feedback enabling manipulation of objects of various sizes and shapes. These works demonstrate the potential of haptic perception and learning techniques for improving the accuracy and efficiency of in-hand manipulation tasks. In practice, tactile sensing provides valuable state information which is hard to extract with alternative methods. 


\subsection{Learning hand transition models}
\label{sec:transition_model}

A common solution for coping with the unavailability of a feasible model is to learn a \textit{transition model} from data. Robot learning problems can typically be formulated as a Markov Decision Process (MDP) \cite{Bellman1957}. Hence, a \textit{transition model}, or \textit{forward model}, is a mapping from a given state $\ve{x}_t\in\ssl{X}$ and action $\ve{a}_t\in\ssl{A}$ to the next state $\ve{x}_{t+1}$, such that $\ve{x}_{t+1} = f(\ve{x}_t,\ve{a}_t)$. Subsets $\ssl{X}$ and $\ssl{A}$ are the state and action spaces of the system, respectively. Such models are commonly obtained through non-linear regression in a high-dimensional space. Often, the forward model is described as a probability distribution function, i.e.,  $P(\ve{x}_{t+1}|\ve{x}_t,\ve{a}_t)$, in order to represent uncertainties in the transition.

Learning transition models for in-hand manipulation tasks typically involves understanding how changes in the robot's state are caused by its actions \cite{nagabandi2020deep}. While a hand transition model is generally available analytically in rigid hands where the kinematics are known \cite{Ozawa2005}, analytical solutions are rarely available for compliant or soft hands. As far as the authors' knowledge, the major work on learning transition models involves such hands. 
Attempts to model compliant hands usually rely on external visual feedback. For example, \citet{Sintov2019} proposed a data-based transition model for in-grasp manipulation with a compliant hand where the state of the hand involves kinesthetic features such as actuator torques and angles along with the position of the manipulated object acquired with visual feedback. The extension of this work used a data-based transition model in an asymptotically optimal motion planning framework in the space of state distributions, i.e., in the belief space \cite{kimmel2019belief,sintov2020motion}. Recently, \citet{morgan2020object} proposed an object-agnostic manipulation using a vision-based Model Predictive Control (MPC) by learning the manipulation model of a compliant hand through an energy-based perspective \cite{morgan2019data}. The work by \citet{Wen2020} used a depth camera to estimate the pose of an object grasped and partly occluded by the two fingers of an underactuated hand. While the work did not consider manipulation, an extension proposed the use of the depth-based 6D pose estimation to control precise manipulation of a grasped object \cite{Morgan2021}. The authors leverage the mechanical compliance of a 3-fingered underactuated hand and utilize an object-agnostic offline model of the hand and the 6D pose tracker using synthetic data. 
While not strictly a transition model, \citet{Calli2018} trained a model to classify transitions and identify specific modes during in-grasp manipulations of an underactuated hand. By using visual and kinesthetic perception, state and future actions are classified to possible modes such as object sliding and potential drop.


While the above methods focus on pure visual perception for object pose estimation, tactile sensors were used in recent work independently or combined with vision. Recent work integrated allocentric visual perception along with four tactile modules, that combine pressure, magnetic, angular velocity and gravity sensors, on two underactuated fingers \cite{Fonseca2019}. These sensors were used to train a pose estimation model. \citet{lambeta2020digit} explored a tactile-based transition model for marble manipulation using a self-supervised detector with auto-encoder architectures. \citet{azulay2022learning} tackled the problem of partial or fully occluded in-hand object pose estimation by using an observation model that maps haptic sensing on an underactuated hand to the pose of the grasped object. Moreover, an MPC approach was proposed to manipulate a grasped object to desired goal positions solely based on the predictions of the observation model. A similar forward model with MPC was proposed by \citet{luo2023} for the multi-finger dexterous Allegro hand. Overall, these approaches demonstrate the potential of using external visual feedback fused with tactile sensing to learn transition models for in-hand manipulation tasks of various hands with uncertainty.


\subsection{Self-supervision and Exploration for Learning Transitions} 

Self-supervision and exploration are important techniques for learning transitions of in-hand robotic manipulation. Self-supervision refers to the process of learning from unlabeled data, where the learning algorithm is able to infer the desired behavior from the structure of the data itself. This can be particularly useful for in-hand manipulation, as it allows the robot to learn about the various states and transitions of an object without the need for explicit human supervision. Exploration, on the other hand, refers to the process of actively seeking out and interacting with the environment in order to collect useful data. In the context of in-hand manipulation, exploration can involve the robot trying out different grasping and manipulation strategies in order to learn what works best for a given object and task. By actively seeking out and interacting with the environment in this way, the robot can learn about the various states and transitions of the object through trial and error, and utilize this knowledge to improve its manipulation performance. Together, self-supervision and exploration can be powerful tools for learning transitions in in-hand manipulation, as they allow the robot to learn from its own experiences and actively gather information about the object being manipulated and its surroundings.

The collection process to generate a state transition model for a robotic system requires active exploration of the high-dimensional state space \cite{kroemer2021}. The common strategy is to exert random actions \cite{calli2018path} in the hope of achieving sufficient and uniform coverage of the robot's state space. In practice, some regions are not frequently visited and consequentially sparse. In systems such as compliant hands \cite{SintovL4DC2020} or object throwing \cite{Zeng2020}, each collection episode starts approximately from the same state and, thus, data is dense around the start state while sparser farther away. Therefore, acquiring state transition models for robotic systems requires exhausting and tedious data collection along with system wear, i.e., the transition function $f(\ve{x}_t,\ve{a}_t)$ is difficult to evaluate.

Active sampling is an alternative strategy where actions that are more informative for a specific task are taken \cite{Wang2018}. However, acquiring a general model of the robot requires the exploration of the entire feasible state space. Bayesian Optimization is the appropriate tool to identify key locations for sampling that would provide increased model accuracy. However, having knowledge of sampling locations does not guarantee the ability to easily reach them. Reaching some state-space regions may require exerting complicated maneuvers. The right actions that will drive the system to these regions for further exploration are usually unknown, particularly in preliminary stages with insufficient data. That is, we require a good model in order to learn a good model.


\section{In-hand Manipulation with Reinforcement Learning}
\label{sec:RL}

\begin{figure*}
    \centering
    \includegraphics[width=\linewidth]{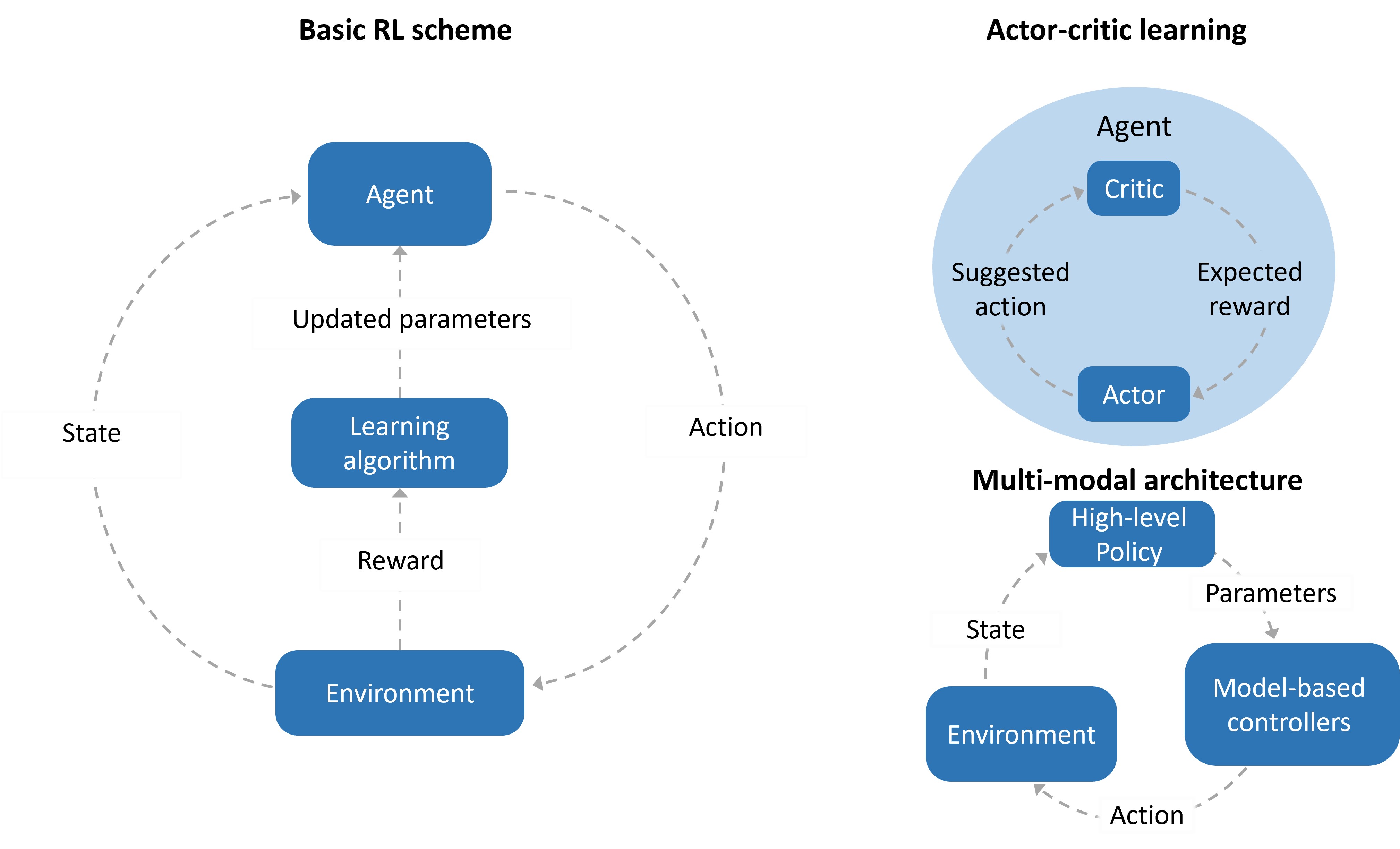} 
    \caption{Illustration of (left) basic RL, (top right) actor-critic architecture, and (bottom right) a multi-network architecture.}
    \label{fig:RL1}
\end{figure*}

Reinforcement learning (RL) is one of the main paradigms of machine learning, akin to supervised and unsupervised learning. RL models learn to take optimal actions within some environments by maximizing a given reward (Figure \ref{fig:RL1}). In contrast to model-driven learning, most RL algorithms collect data during the learning process. Often, the learning is done in simulated environments in order to avoid tedious work and wear of the real robot. RL policies are functions mapping current states to optimal actions and a distinction is commonly made between on- and off-policy learning \cite{Singh2022}. Both approaches commonly approximate the value function which is an expected cumulative reward defined for states and actions. In on-policy learning, data collection is guided by the intermediate policy learned by the agent. The value function is learned directly by the policy. 
Therefore, a balance must be kept between exploration of unvisited action-state regions, and exploitation of known regions in order to maximize reward. In off-policy methods, on the other hand, the value function of the optimal policy is learned independent of the actions conducted by the agent during training. 

\subsection{Brief RL overview} 

Assuming that a given system is a Markov Decision Process (MDP), the next state depends solely on the current state and desired action according to a forward transition dynamics while receiving a reward. In model-based RL, a transition model can be learned independently of the policy learning as described in Section \ref{sec:transition_model}. When the system is stochastic due to uncertainties and limited observability, a Partial Observed Markov Decision Process (POMDP) is considered. As the true state cannot be fully observed in such a case, an observation space is introduced. Therefore, the agent receives an observation when reaching the next state with some probability. In both MDP and POMDP, the goal is, therefore, to learn a policy which maximizes the expected reward. A wider review of key concepts and methods can be found in the work of \citet{Nguyen2019}.

Deep reinforcement learning integrates the RL learning paradigm with deep ANNs serving as the policy or value function approximation. Such integration has revealed significant capabilities and led to the success of many reinforcement learning domains \cite{mnih2013playing,mnih2015human} including robot manipulation \cite{levine2016end,Nguyen2019} and specifically in-hand dexterous manipulation \cite{Jain2019,andrychowicz2020learning}. Examples of straight-forward implementation of an RL algorithm include the work of \citet{antonova2017reinforcement} which addressed the pivoting problem with a parallel gripper. The RL policy was trained while relying on fast tracking with a camera. However, the trained policy yielded excessive back and forth motions. \citet{Cruciani2017} coped with this limitation by employing a three-stage manipulation in which the robot learns to control the velocity and opening of the gripper. An RL policy was acquired through Q-Learning \cite{watkins1992q}. In an extension work, \citet{Cruciani2018} integrated path planning with the RL policy for the robotic arm to perform more complex tasks. In another example, \citet{VanHoof2015} was the first to employ RL on a two-finger underactuated hand by utilizing its compliance and tactile sensing. 

Direct implementation of RL algorithms usually only works in specific and limited applications. Hence, this section presents a comprehensive survey of advanced approaches and current research in the field, with a focus on the unique challenges of in-hand manipulation. First, we describe transfer learning challenges and approaches focusing on the \sr problem. Next, the problem of episodic resetting in real-life experiments is discussed. Finally, we explore the topic of multi-level control systems and actor-critic learning schemes.  


\subsection{Transfer learning and \sr problems}

Often, ANN based controllers must train extensively for each new task before being able to perform successfully, requiring long training periods and extensive computation resources. Specifically for in-hand manipulation, performing tasks with trained policies on new objects may be challenging. Transfer learning can be divided into few-shot, one-shot and zero-shot learning \cite{Pourpanah2023}. Few-shot and one-shot transfer learning requires few instances or a single instance, respectively, of a new task in order to tune the previously trained model. On the other hand, a trained model in zero-shot transfer can instantly perform tasks not included in the training stage. Thus, the ability to learn and generalize to new tasks in few-shots or less is highly beneficial. In in-hand manipulation, the transfer of a model often refers to the generalization to new objects not included in the training \cite{Huang2021}.

The common approach in few- and one-shot transfer learning is to share weights and data between different tasks, objects and hands. \citet{funabashi2019morphology} demonstrated the ability to pre-train a policy to perform a stable rolling motion with only three fingers of the Allegro hand and then transfer to utilize all four fingers. This was shown to be possible given identical finger morphologies. It was shown that pre-training can be done with data gathered even from random motions such that, afterward, training for specific tasks can be done in one-shot transfer. 

While training RL models on real robots yields highly successful controllers \cite{kalashnikov2018scalable,zhu2019dexterous}, it is also expensive in time and resources, or can pose danger with some robots. Furthermore, it tends to require extensive human involvement as discussed in the next section.
Consequently, simulations are an emerging approach for policy training as they enable rapid and efficient collection of a massive amount of data. While training on simulations can be beneficial, transferring a robot policy trained in simulation to a real robot remains a challenge \cite{zhao2020sim}. Compared to real-world systems that are usually uncertain and noisy, simulations are naturally more certain and simplified. This gap is commonly known as the \sr problem and can significantly reduce the performance of policies trained in the simulation domain and transferred to the application domain in the real world \cite{hofer2021sim2real}. This is especially relevant in the case of in-hand manipulation tasks which tend to heavily involve hard-to-model contact dynamics \cite{funabashi2020variable,liarokapis2016post}. Hence, the resulting controllers are often sensitive to small errors and external disturbances. 

The most common approach for bridging the reality gap in a \sr problem is \textit{domain randomization} \cite{van2019sim}. In this approach, various system parameters in the simulation are constantly varied in order to improve robustness to modeling errors. \citet{OpenAI2019,andrychowicz2020learning} proposed the Automatic Domain Randomization (ADR) approach where models are trained only in simulation and can be used to solve real-world robot manipulation problems. Specifically, a Rubik's cube was solved by performing finger gaiting and rolling manipulations with the anthropomorphic Shadow hand. ADR automatically generates a distribution over randomized environments. Control policies and vision state estimators trained with ADR exhibit vastly improved \sr transfer. 

In the work of \citet{Sievers2022}, a PyBullet simulation of the DLR hand was used to train an off-policy for in-grasp manipulation solely using tactile and kinesthetic sensing. Domain randomization was used for \sr transfer to the real hand. An extension of the work has demonstrated zero-shot \sr transfer while focusing on 24 goal orientations \cite{Pitz2023}. Beyond directly modifying the dynamics in domain randomization, applying small random forces to the grasped object was shown by \citet{allshire2022transferring} to improve the robustness of the resulting policy in in-grasp manipulation of the TriFinger hand. Recently, \citet{Handa2023} have taken the domain randomization approach to reorient a cube within the four-finger Allegro hand using RGB-D perception. As opposed to Pybullet and similar simulators which are based on CPU computations, \citeauthor{allshire2022transferring} and \citeauthor{Handa2023} used the GPU-based Nvidia Isaac Gym simulator \cite{makoviychuk2021isaac}. Using a GPU-based simulator reduces the amount of computational resources and costs. 

As opposed to domain randomization, \citet{qi2022} used an adaptation module learning to cope with the \sr problem. The module trained through supervised learning to approximate the important properties of the system based solely on kinesthetic sensing. An RL policy is trained to take actions for finger gaiting with a multi-finger hand based on the approximations and on real-time state observations. An extension of the work added visual and tactile perception while also including a Transformer model for embedding past signals \cite{Qi2023}. \citeauthor{Qi2023} have also used the Isaac Gym simulator since it excels in contact modeling. However, Isaac Gym and most other simulators tend to provide unreliable contact force values. To cope with this limitation, \citet{Yin2023} simulated 16 tactile sensors across a four-finger Allegro (i.e., fingertips, fingers and palm) while considering only binary signals of contact or no contact. Due to this configuration, a trained policy is shown to successfully ease the \sr transfer.




\subsection{Episodic resetting}

Learning robot tasks in the real world often requires sufficient experience. In many systems, this is commonly achieved with frequent human intervention for resetting the environment in between repeating episodes, for example when the manipulated object is accidentally dropped. It is particularly relevant in the case of in-hand manipulation where resetting may be more complex due to large uncertainties in failure outcomes. Removing the costly human intervention will improve sample collection and, thus, decrease learning time. \citet{eysenbach2018leave} proposed a general approach for training a reset policy simultaneously with the task policy. For instance, a robot manipulator can be trained to reset the environment within the policy training allowing a more autonomous and continuous learning process. As shown by \citet{srinivasan2020learning}, the resulting reset policy can be used as a critic for the task controller in order to discern unsafe task actions that will lead to irreversible states, where reset is inevitable. Specifically in this work, a model learns to identify actions that a Shadow hand may exert while attempting to reorient a cube through rolling and finger gaiting, without the risk of dropping the cube entirely. Preventing the reach of these irreversible states increases the safety of the controller, and can also be used to induce a curriculum for the forward controller. 

Another approach to avoid irreversible states is by the addition of a reactive controller designed specifically for intervening only when the robot state is in the close neighborhood of such irreversible states \cite{Falco2018}. In this work, \citeauthor{Falco2018} used a compliant prosthetic hand in the in-grasp manipulation of objects based on visual perception with an added reactive controller connected to tactile sensors. The goal of the reactive controller is to avoid object slipping. The nominal control method can, thus, be trained with the goal of not only succeeding in the given task but also minimizing the intervention of the reactive controller.

While episodic resetting is often considered a burden, it can instead be considered an opportunity. When training multi-task capabilities for in-hand manipulation, failure in one task may cause a need for a resetting of the grasp. Rather than using human intervention or an additional control system, the reset can instead be viewed as another manipulation task \cite{Gupta2021}. For example, an unsuccessful attempt at a rolling motion 
which leads to a wrong object pose, may require the learning of a sliding task to fix the pose. Thus, task training ending in success or failure can both be chained to further learning of other tasks. This results in a reset-free learning scheme.


\subsection{Multi-network architecture} 

Multi-network architectures, such as actor-critic \cite{lillicrap2015continuous} or teacher-student \cite{zimmer2014teacher}, are often beneficial in improving the learning process. In the more common actor-critic structure, an actor network is trained as the policy while the critic network is trained to estimate the value function. Such structures try to cope with the inherent weaknesses of single-network structures. That is, actor-only models tend to yield high variance and convergence issues while critic-only models have a discretized action space and, therefore, cannot converge to the true optimal policy. In teacher-student architectures, on the other hand, knowledge distillation enables the transfer of knowledge from an unwieldy and complex model to a smaller one. As such, a teacher model is an expert agent that has already learned to take optimal actions whereas the student model is a novice agent learning to make optimal decisions with the guidance of the teacher. 

\citet{chen2021simple, chen2022,Chen2023} used an asymmetric teacher-student training scheme with a teacher trained on full and privileged state information. Then, the teacher policy is distilled into a student policy which acts only based on limited and realistic available information. Policies for object reorientation tasks were trained with a simulated Shadow hand on either the EGAD \cite{morrison2020egad} or the YCB \cite{calli2015ycb} benchmark object sets and tested on the other. Results have exhibited zero-shot transfer to new objects. While the teacher-student approach utilizes privileged information during training, the actor-critic approach manages the learning by continuous interaction between the two models. This was demonstrated with the Proximal Policy Optimization (PPO) algorithm for in-hand pivoting of a rigid body held in a parallel gripper, using inertial forces to facilitate the relative motion \cite{Toledo2021}. Moreover, combining actor-critic methods with model-based methods can result in improved learning. The learned model can be used within a model predictive controller to reduce model bias induced by the collected data. This was demonstrated by an underactuated hand to perform finger gaiting \cite{morgan2021model} and object insertion \cite{azulay2022haptic}. Recently, \citet{Tao2023} proposed to consider the multi-finger hand during a reorientation task as a multi-agent system where each finger or palm is an agent. Each agent has an actor-critic architecture while only the critic has a global observation of all agents. The actor, on the other hand, has only local observability of neighboring agents. In such a way, the hand does not have centralized control and can adapt to changes or malfunctions.

In a different multi-modal architecture approach proposed by \citet{Li2020}, various control tasks required for in-hand manipulation are divided into multiple hierarchical control levels (Figure \ref{fig:RL1}). This allows the use of more specialized tools for each task. In the lower level, traditional model-based controllers robustly execute different manipulation primitives. On the higher level, a learned policy orchestrates between these primitives for a three-finger hand to robustly reorient grasped objects in a planar environment with gravity. A similar approach was taken by \citet{Veiga2020} where low-level controllers maintain a stable grasp using tactile feedback. At the higher level, an RL is trained to perform in-grasp manipulation with a multi-finger dexterous hand.


\subsection{Curriculum Learning}

Often, directly training models with data from the entire distribution may yield insufficient performance. Hence, Curriculum Learning (CL) is a training strategy where the model is gradually exposed to increasing task difficulty for enhanced learning efficacy \cite{wang2021survey}. Such a process imitates the meaningful learning order in human curricula. Adding CL to guide the development of necessary skills can aid policies to learn difficult tasks that tend to have high rates of failure \cite{chen2022}. In this example, researchers modify the behavior of gravity in simulations according to the success rate to aid the learning of gravity-dependent manipulation tasks. This method allows the robot to first successfully learn a skill and then move on to increasingly harder and more accurate problems, slowly reaching the actual desired skill. \citet{azulay2022haptic} trained an actor-critic model to insert objects into shaped holes while performing in-grasp manipulations with a compliant hand. The work has exhibited object-based CL where simple objects were first introduced to the robot followed by more complex ones. Other uses of CL can involve guiding the exploration stages of solution search using reward shaping \cite{allshire2022transferring}. The work shows that it is possible to improve early exploration by guiding the model directly to specific regions using specific reward functions as priors. Those regions however may not hold actual feasible solutions, and it may be necessary to reduce the effects of the reward functions in later stages of the learning process. 


\subsection{Tactile information}

While visual perception is a prominent approach for feedback in RL, it may be quite limited in various environments and the object is often occluded by the hand. On the other hand, tactile sensing provides direct access to information regarding the state of the object. Nevertheless, data from tactile sensing is often ambiguous, and information regarding the object is implicit. Yet, the addition of tactile sensory is widely addressed as it can improve the learning rate. For instance, in the work of \citet{korthals2019multisensory}, tactile sensory information for the Shadow hand increased the sampling efficiency and accelerated the learning process such that the number of epochs for similar performance was significantly decreased. 
\citet{Jain2019} have shown that the integration of tactile sensors increases the learning rate when the object is highly occluded. This was demonstrated in various manipulation tasks including in-hand manipulation of a pen by a simulated anthropomorphic hand. \citet{melnik2019tactile} compared multiple sensory methods, including continuous versus binary (i.e., touch or no touch) tactile signals and, higher versus lower sensory resolutions. The results from this comparative study have shown that using tactile information is beneficial to the learning process, compared to not having such information. However, the specific method that gave the best result was dependent on the learned manipulation task \cite{melnik2021}. 

While \citeauthor{melnik2021} used tactile information directly as part of the state vector, \citet{funabashi2020variable,funabashi2020stable} used a higher-resolution sensory array without visual perception. The model coped with the increased dimension of the output tactile information by considering the relative spatial positioning of the sensors. Similarly, \citet{Yang2023} used a tactile array across a multi-finger hand. The array was embedded using Graph Neural Network which provides an object state during the manipulation and used for model-free RL. Recently, \citet{Khandate2022} implemented model-free RL to reorient an object through finger gaiting with a multi-finger dexterous hand while only using kinesthetic and tactile sensing. In an extension work, \citeauthor{Khandate2022} offered the use of sampling-based motion planning in order to sample useful parts of the manipulation space and improve the exploration \cite{Khandate2023}. Tactile sensing provides valuable information and often can fully replace visual perception. However, policies based on tactile perception usually require an excessive amount of real-world experience in order to reach sufficient and generalized performance.

\hfill

In summary, research in the field of RL for robotic in-hand manipulation is growing and achieving increasing success in recent years.  While showcasing promising performance in specific tasks, RL policies still perform poorly in multi-task scenarios and struggle to generalize with zero- or few-shot learning \cite{Chen2023bi}. Major challenges currently being faced include the control of highly dexterous hands with high amounts of sensory information, the transfer of a model learned in simulation to a real robotic system, and the transfer of learning specific tasks on well-known objects to other tasks and unknown objects. Major advances are being achieved using multi-level control structures, domain and dynamic adaptation, and the combination of model-based and model-free methods to gain the benefits of both. While exciting advances have been made in RL, the field continues to explore the challenges of data efficiency and adaptability to new domains.





\section{Imitation Learning for In-hand Manipulation}


As discussed in the previous section, training RL policies for real robots from scratch is usually time-consuming and often infeasible due to the lack of sufficient data \cite{rapisarda2019reinforcement}. A prominent approach for coping with these challenges is Imitation Learning (IL). Instead of learning a skill without prior knowledge, IL aims to learn from expert demonstrations \cite{Duan2017,fang2019survey}. Prior knowledge from the expert can then be optimized for the agent through some learning framework such as RL. While IL is often considered a sub-field of RL, we provide a distinct focus due to its importance and wide work. IL can be categorized into two main approaches: Behavioral Cloning (BC) and Inverse Reinforcement Learning (IRL) 
\cite{zheng2022imitation}. In BC, a policy is trained in a supervised learning fashion with expert data to map states to actions. IRL, on the other hand, extracts the reward function from the expert data in order to train an agent with the same preferences \cite{Arora2021}. 

\begin{figure}[]
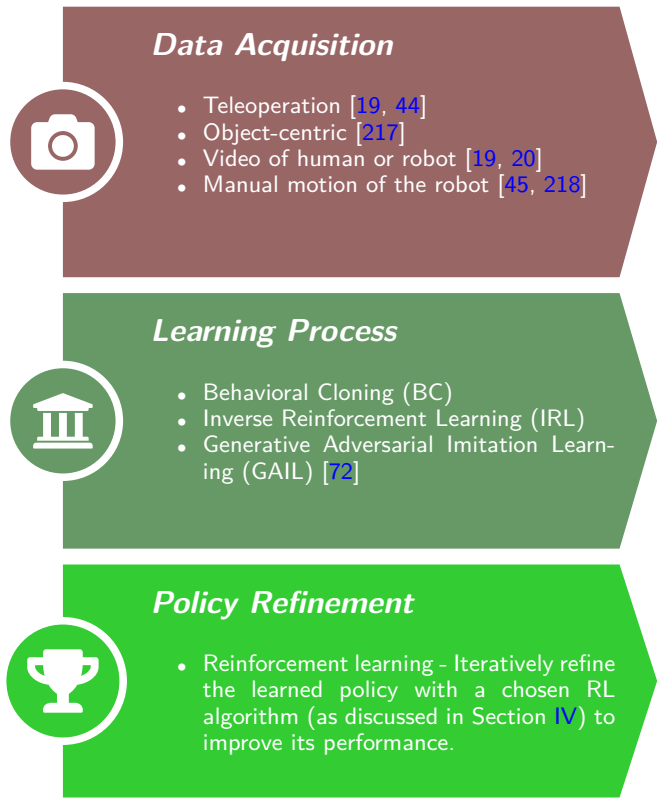

\centering
\begin{mybox}{Data Acquisition}{\faIcon[regular]{camera}}{gray!80!red}{3.6cm}
\begin{itemize}
    \item Teleoperation \cite{Arunachalam2023,Kumar2016b}
    \item Object-centric \cite{Gupta2016}
    \item Video of human or robot \cite{Arunachalam2023, Arunachalam2023holo}
    \item Manual motion of the robot \cite{Li2014,Gaspar2018}
\end{itemize}
\end{mybox}
\begin{mybox}{Learning Process}{\faIcon[regular]{university}}{gray!80!green}{3.4cm}
\begin{itemize}
    \item Behavioral Cloning (BC)
    \item Inverse Reinforcement Learning (IRL)
    \item Generative Adversarial Imitation Learning (GAIL) \cite{Wei2023}
\end{itemize}
\end{mybox}
\begin{mybox}{Policy Refinement}{\faIcon[regular]{trophy}}{green!60!gray}{3.1cm}
\begin{itemize}
    \item Reinforcement learning - Iteratively refine the learned policy with a chosen RL algorithm (as discussed in Section \ref{sec:RL}) to improve its performance.
\end{itemize}
\end{mybox}
\vspace{-0.02in}
\caption{Flowchart of policy training with Imitation Learning (IL). The  policy is first learned based on expert demonstrations and then iteratively refined using a chosen RL algorithm.}
\label{fig:IL_FC}
\end{figure}

In both BC and IRL, a policy is learned with some prior. This is in contrast to RL where the policy is learned from scratch based on the agent's own experience. Hence, IL requires an initial process of data acquisition as illustrated in Figure \ref{fig:IL_FC}. First, data is collected from demonstrations of an expert. Demonstrations can be acquired in various mediums such as recording human motion or recording proprioceptive sensing of the robot during manual jogging. In the next step, IL usually involves either learning a policy to directly imitate the demonstration (BC) or feature extraction from the data (IRL). The last step is further policy refinement through conventional RL. From an RL perspective, IL usually reduces the learning time by bootstrapping the learning process using an approximation of the expert's policy. 

\subsection{Date Acquisition}

Data is collected from an expert demonstrator while conducting the desired task. The motion of the expert is recorded through some set of sensors such that the learning agent can later observe and learn to imitate. There are various approaches to demonstrate and record the motions and their choice may affect the learning process. 

One data acquisition approach is to teleoperate the robot throughout the task using designated tools such as a remote control \cite{zhang2018deep}. However, remote controls are unnatural and quite infeasible in teleoperation of dexterous robotic hands. In a more natural approach, \citet{Arunachalam2023} used a visual hand pose estimation model (i.e., skeleton) to approximate keypoints on the human hand during reorientation of an object. The user can also use a VR set in order to have the point-of-view of the robot \cite{Arunachalam2023holo}. In these examples, a policy is learned for an anthropomorphic robot hand by using simpler nearest-neighbors search in the data. The action in the demonstration data which has a state closest to the current state is exerted. Similarly, \citet{Kumar2016b} recorded the proprioceptive state of a virtual anthropomorphic robotic hand during teleoperation with the CyberGlove worn by an expert user. With the glove, the joint angles along with tactile information are recorded. The recorded tasks are then used to train and evaluate in-hand manipulation with a five-finger dexterous hand for reorientation tasks. In a similar approach, \citet{wei2023wearable} designed a wearable robotic hand for IL teleoperation such that the expert has tactile feedback during demonstrations.

In a different approach by \citet{Gupta2016}, only information regarding the motion of a manipulated object is collected while ignoring the motions of the human expert. Hence, an object-centric policy is learned while selecting the most relevant demonstration for each initial state in the training. In a different approach, the demonstrator manually moves the robot by contacting and pushing it to perform the task \cite{Li2014,Shin2024,Gaspar2018}. During the demonstration, the robot collects kinesthetic data from the joints. While the approach is simple, it is usually applied to robotic arms with a single serial kinematic chain. It is quite infeasible for a human to synchronously move a dexterous and multi-contact robotic hand to perform a complex in-hand manipulation task. Nevertheless, simpler tasks with non-dexterous hands may by possible while the authors have not found prior work.

\subsection{Learning Process}

The process for learning from the demonstrations is commonly conducted by either BC or IRL \cite{hussein2017imitation}. In BC, the agent is required to directly take the strategy of the expert observed in the demonstrations \cite{Arunachalam2023}. The agent will exert an action taken by the expert when in a similar state. Hence, demonstration data is usually recorded in the form of state-action pairs which is easy to learn. Then, a policy is learned in a supervised learning manner. However, state-action pairs can be difficult to obtain from, for instance, video data. To cope with this problem, \citet{radosavovic2020state} proposed the state-only imitation learning (SOIL) approach where an inverse dynamics model can be trained to extract the actions chosen based on the change in the state perceived from videos. The inverse dynamics model and the policy are trained jointly. SOIL enables learning from demonstrations originating from different but related settings. While not an IL approach, \citet{Yuan2023} considered a trained teacher policy as an expert and used BC to distill it to a student in the training of in-hand manipulation with vision and tactile sensing. In a different work, BC was used to control a unique design of a gripper having actuated rollers on its fingertips \cite{Yuan2020}. The demonstration data was extracted from a handcrafted controller and shown to improve performance.

\citet{Rajeswaran2017} compared methods of RL to solve complex manipulation tasks, with and without incorporating human demonstrations. The authors suggested a method of incorporating demonstrations into policy gradient methods for manipulation tasks. The proposed Demonstration Augmented Policy Gradient (DAPG) method uses pre-training with BC to initialize the policy and an augmented loss function to reduce ongoing bias toward the demonstration. The results in the paper showcase that DAPG policies can acquire more human-like motion compared to RL from scratch and are substantially more robust. In addition, the learning process is considerably more sample-efficient. \citet{Jain2019} extended the work by exploring the contribution of demonstration data to visuomotor policies while being agnostic about the data's origin. Demonstrations were shown to improve the learning rate of these policies in which they can be trained efficiently with a few hundred expert demonstration trajectories. In addition, tactile sensing was found to enable faster convergence and better asymptotic performance for tasks with a high degree of occlusions. While \citeauthor{Rajeswaran2017} and \citeauthor{Jain2019} demonstrated the approach only in simulations, \citet{zhu2019dexterous} demonstrated the use of DAPG on a real-robot in complex dexterous manipulation tasks. The results have shown a decrease of training time from 4-7 hours to 2-3 hours by incorporating human demonstrations. 

Few studies on in-hand manipulation have used BC due to the significant effort required to collect sufficient demonstration data. While simple to implement, BC usually requires large amounts of data for sufficient performance \cite{Ross2011}. IRL, on the other hand, directly learns the reward function of the demonstrated expert policy which prioritizes some actions over others \cite{Arora2021}. IRL learns the underlying reward function of the expert which is the best definition of a task. Once acquired the reward function, an optimal policy can be trained to maximize such a reward using a standard RL algorithm. While general work on IRL is wide for various robotic applications, not much work has been done that combines IRL with in-hand manipulation. A single work demonstrated the IRL approximation of the reward function using expert samples of desired behaviours \cite{Orbik2021}. However, the authors have argued that the learned reward functions are biased towards the demonstrated actions and fail to generalize. Randomization and normalization were used to minimize the bias and enable generalization between different tasks. 

While not directly IRL, \citet{deng2020learning} utilized reward-shaping to improve the RL training of in-hand manipulation with a dexterous hand. By observing hand synergies of a human demonstrator, a limited and low-dimensional state space was constructed. Using reward-shaping allows the inclusion of multiple levels of knowledge, from the standard extrinsic reward to hand synergies-based reward and an uncertainty-based reward function that is aimed at directing efficient exploration of the state space. Learning using all three reward functions is shown through simulations to improve learning. The minor use of IRL to address in-hand manipulation problems may be explained by its tendency to provide ill-behaved reward functions and unstable policies \cite{cai2019global}.

IL was also proposed for in-hand manipulation without the use of RL. \citet{Solak2019} proposed the use of Dynamical Movement Primitives (DMP) \cite{Ijspeert2013}. The approach shows that a multi-finger dexterous hand can perform a task based on a single human demonstration while being robust to changes in the initial or final state, and is object-agnostic. However, the former property may yield object slip and compromise grasp stability. Hence, an extended work proposed haptic exploration of the object such that the manipulation is informed by surface normals and friction at the contacts \cite{Solak2023}.

While traditional IRL has shown high performance in a wide range of tasks, it only provides a reward function that implicitly explains the experts' behaviour but does not provide the policy dictating what actions to take. Hence, the agent will still have to learn a policy through RL training in a rather expensive process. To address this problem, the Generative Adversarial Imitation Learning (GAIL) \cite{Ho2016} was proposed and combines IL with Generative Adversarial Networks (GAN) \cite{Goodfellow2020}. Similar to GAN, GAIL incorporates a generator and a discriminator. While the generator attempts to generate a policy that matches the demonstrations, the discriminator attempts to distinguish between data from the generator and the original demonstration data. Training of GAIL is, therefore, the minimization of the difference between the two. Consequently, GAIL is able to extract a policy from the demonstration data. Recently, the use of GAIL was proposed for in-hand manipulation by a dexterous hand \cite{Wei2023}. The approach was shown to perform significantly better than BC or direct RL training. GAIL has the potential to improve and expedite policy learning of more complex in-hand manipulation tasks, and should be further explored.


\begin{table*}
    \centering
    \caption{Comparison of key components in learning methods for robotic in-hand manipulation.}
    \label{tb:comparison}
    \includegraphics[width=\linewidth]{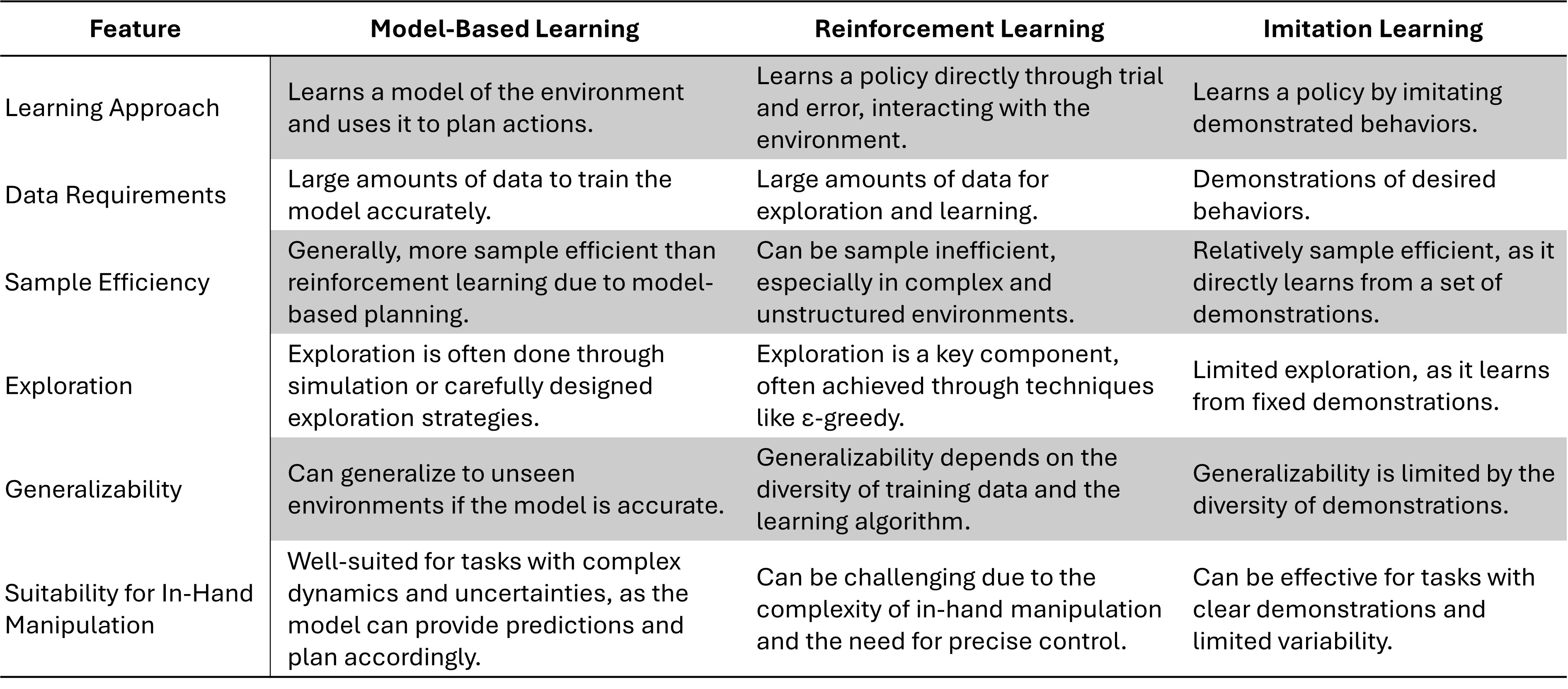} 
\end{table*}

\section{Discussion}
\label{sec:discussion}


In-hand manipulation is one of the most challenging topics in robotics and an important aspect for feasible robotic applications. Traditional analytical methods struggle to estimate object properties and noisy sensory information. With in-hand manipulation reaching a bottleneck using these traditional methods, researchers are leveraging advancements in deep learning and reinforcement learning to unlock new levels of dexterity. A summary comparison of the three learning approaches discussed in this paper is given in Table \ref{tb:comparison}. These tools encapsulate the ability to model complex and noisy systems such as a dexterous robotic hand equipped with various sensors. Nevertheless, current research still faces significant challenges:

\begin{enumerate}
    \item \textit{Data efficiency.} Learning models is essential for understanding changes in the robot's state caused by its actions during in-hand manipulation. While analytical solutions are available for rigid hands, compliant or soft hands rely on external visual feedback. However, collecting data can be challenging due to the high-dimensional state space and the need to explore the entire feasible space. Future work by researchers should address methods to reduce the required size of training data by making models more general to various applications. For instance, Bayesian optimization can assist in identifying key sampling locations, but reaching some regions may require complex maneuvers, making it necessary to have a good prior model to learn a better one.
    
    \item \textit{Sim-to-real transfer.} Learning policies in simulation is a prominent approach to improve data efficiency in robot training. While significant progress has been made to address the \sr problem, simulations hardly represent the real world and trained policies work poorly on the real system. Hence, large efforts should be put into closing the reality gap by generating better simulations and, incorporating advanced data-based models that can generalize better. Examples of the latter include decision transformers \cite{Monastirsky2023} and diffusion policies \cite{chi2023diffusionpolicy}. These advanced methods are versatile and can be applied to either of the three learning paradigms:  model-based learning, RL and IL.

    \item \textit{Soft robotic hands.} High-dexterous hands such as anthropomorphic ones have been demonstrated in multiple complex in-hand manipulation tasks. However, they are highly expensive making their adaptation to real-world tasks not possible. Consequently, an abundance of research and development has been put in recent years on soft robotic hands that are typically low-cost to manufacture. However, these hands cannot be modeled or controlled analytically and learning approaches are the common paradigm. As discussed previously, common solutions require a significant amount of data and are usually specific to a single hand and task. Therefore, the robotics community should promote efficient learning approaches in terms of data efficiency, computationally light-weight and generalizable to different hardware, tasks and environments. Specifically, future research should prioritize the development of more realistic simulation environments tailored for soft and adaptive robotic hands.
    
    \item \textit{Tactile sensing.} While visual perception technology is quite mature, the use of high-resolution tactile sensing is relatively new. In general, Table \ref{tb:Pcomparison} clearly shows the dominance of visual perception over tactile sensing in research. Highly capable tactile sensors can provide vital information regarding the contact state including position, forces, torsion, shape and texture. Nevertheless, they often require a large amount of real-world data in order to perform well. Simulations such as TACTO \cite{wang2022tacto} address this problem by simulating tactile interactions. However, these remain quite far from reality and cannot provide reliable load sensing. Practitioners should work toward better tactile simulators along with distillation approaches for efficient \sr transfer.

    \item \textit{Learning from Demonstrations.} IL with expert demonstrations has proved to be efficient for shortening the data-hungry training phase of RL. However, hardware and methods for collecting demonstration data generally lack the ability to capture the entire state space of the hand-object system. For instance, visual perception is incapable of observing the intrinsic and contact state of the system. Furthermore, IL models focus on task completion and fail to address strategy learning with efficient data utilization. Future work should facilitate efficient platforms for collecting high-dimensional data in the real world. In addition, learning methods should require a small amount of data from the expert user in order to generalize well to various scenarios of the tasks. 

    \item \textit{Task generalization.} The prevailing paradigm in in-hand manipulation focuses on crafting task-specific or narrowly applicable policies, which hinders broader applicability. Collected datasets typically consist of several tens of thousands of samples tailored to the specific task at hand. The field therefore necessitates a paradigm shift toward solutions capable of seamless adaptation or generalization to novel tasks or objects. A large, standard and unified dataset of in-hand manipulation in-the-wild assembled by many researchers would be invaluable for advancing generalization.
\end{enumerate}



\section{Conclusions}

This paper provides a comprehensive survey of various learning-based approaches for robotic in-hand manipulation, focusing on model-based methods, reinforcement learning (RL), and imitation learning (IL). Each of these methodologies has demonstrated significant progress in enabling robotic systems to perform dexterous in-hand manipulation tasks, which are essential for robots to operate effectively in complex human environments. Despite these advancements, several challenges remain, such as the need for higher data efficiency, improved sim-to-real transfer and better generalization across different objects and tasks.

While RL has revealed success due to its ability to generate solutions with minimal human intervention. Key findings indicate that RL policies often struggle with generalization and multi-task scenarios. Similarly, model-based approaches offer precision but can be limited by the complexity of dynamic environments. Imitation learning provides a promising avenue by leveraging expert demonstrations, but it requires extensive data collection, and its performance is highly dependent on the quality of the demonstrations. In addition to the challenges and future research suggestions discussed in Section \ref{sec:discussion}, advancements should also be made in more applicative directions such as: enhance the generalization of models to be agnostic to the robotic hand with versatility to various tasks, through few-shot or zero-shot learning; augment the capabilities of prosthetic hands to perform more complex tasks that usually involve in-hand manipulation; explore simplistic multimodal sensing while efficiently integrating these modalities; and, utilize the significant potential in human demonstration and continuous learning during human-robot collaboration, where robots can learn from human demonstrations and adapt to human preferences. By addressing these challenges, future research can push the boundaries of robotic dexterity, enabling robots to perform more sophisticated tasks autonomously.


\printbibliography

@ARTICLE{Azulay2024,
  author={Azulay, Osher and Curtis, Nimrod and Sokolovsky, Rotem and Levitski, Guy and Slomovik, Daniel and Lilling, Guy and Sintov, Avishai},
  journal={IEEE Robotics and Automation Letters}, 
  title={AllSight: A Low-Cost and High-Resolution Round Tactile Sensor With Zero-Shot Learning Capability}, 
  year={2024},
  volume={9},
  number={1},
  pages={483-490}}

@article{LlopHarillo2019,
title = {The Anthropomorphic Hand Assessment Protocol (AHAP)},
journal = {Robotics and Autonomous Systems},
volume = {121},
pages = {103259},
year = {2019},
doi = {https://doi.org/10.1016/j.robot.2019.103259},
url = {https://www.sciencedirect.com/science/article/pii/S0921889019300946},
author = {Immaculada Llop-Harillo and Antonio Pérez-González and Julia Starke and Tamim Asfour},
}

@article{fang2019survey,
  title={Survey of imitation learning for robotic manipulation},
  author={Fang, Bin and Jia, Shidong and Guo, Di and Xu, Muhua and Wen, Shuhuan and Sun, Fuchun},
  journal={International Journal of Intelligent Robotics and Applications},
  volume={3},
  number={4},
  pages={362--369},
  year={2019},
  publisher={Springer}
}

@inproceedings{zhang2018deep,
  title={Deep Imitation Learning for Complex Manipulation Tasks from Virtual Reality Teleoperation},
  author={Zhang, Tianhao and McCarthy, Zoe and Jow, Owen and Lee, Dennis and Chen, Xi and Goldberg, Ken and Abbeel, Pieter},
  booktitle={IEEE International Conference on Robotics and Automation (ICRA)},
  pages={5628--5635},
  year={2018},
}

@article{Arora2021,
title = {A survey of inverse reinforcement learning: Challenges, methods and progress},
journal = {Artificial Intelligence},
volume = {297},
pages = {103500},
year = {2021},
doi = {https://doi.org/10.1016/j.artint.2021.103500},
author = {Saurabh Arora and Prashant Doshi},
}

@article{Park2024,
author = {Park, Wookeun and Park, Seongjin and An, Hail and Seong, Minho and Bae, Joonbum and Jeong, Hoon Eui},
title = {A Sensorized Soft Robotic Hand with Adhesive Fingertips for Multimode Grasping and Manipulation},
journal = {Soft Robotics},
pages = {null},
year = {2024},
doi = {10.1089/soro.2023.0099},
}

@Article{Shin2024,
AUTHOR = {Shin, Ku Jin and Jeon, Soo},
TITLE = {Nonprehensile Manipulation for Rapid Object Spinning via Multisensory Learning from Demonstration},
JOURNAL = {Sensors},
VOLUME = {24},
YEAR = {2024},
NUMBER = {2},
ARTICLE-NUMBER = {380},
DOI = {10.3390/s24020380}
}

@INPROCEEDINGS{GarciaHernando2020,
  author={Garcia-Hernando, Guillermo and Johns, Edward and Kim, Tae-Kyun},
  booktitle={IEEE/RSJ International Conference on Intelligent Robots and Systems (IROS)}, 
  title={Physics-Based Dexterous Manipulations with Estimated Hand Poses and Residual Reinforcement Learning}, 
  year={2020},
  volume={},
  number={},
  pages={9561-9568},
  doi={10.1109/IROS45743.2020.9340947}}

@inproceedings{Olson2011,
    TITLE      = {{AprilTag}: A robust and flexible visual fiducial system},
    AUTHOR     = {Edwin Olson},
    BOOKTITLE  = {{IEEE} International Conference on Robotics and
                 Automation ({ICRA})},
    YEAR       = {2011},
    PAGES      = {3400-3407},
}

@article{GarridoJurado2014,
title = {Automatic generation and detection of highly reliable fiducial markers under occlusion},
journal = {Pattern Recognition},
volume = {47},
number = {6},
pages = {2280-2292},
year = {2014},
author = {S. Garrido-Jurado and R. Muñoz-Salinas and F.J. Madrid-Cuevas and M.J. Marín-Jiménez},
}

@ARTICLE{Billings2019,
  author={Billings, Gideon and Johnson-Roberson, Matthew},
  journal={IEEE Robotics and Automation Letters}, 
  title={SilhoNet: An RGB Method for 6D Object Pose Estimation}, 
  year={2019},
  volume={4},
  number={4},
  pages={3727-3734},
  doi={10.1109/LRA.2019.2928776}}

@INPROCEEDINGS{Rad2017,
  author={Rad, Mahdi and Lepetit, Vincent},
  booktitle={IEEE International Conference on Computer Vision (ICCV)}, 
  title={BB8: A Scalable, Accurate, Robust to Partial Occlusion Method for Predicting the 3D Poses of Challenging Objects without Using Depth}, 
  year={2017},
  volume={},
  number={},
  pages={3848-3856},
  doi={10.1109/ICCV.2017.413}}

@article{Kalaitzakis2021,
author = {Kalaitzakis, Michail and Cain, Brennan and Carroll, Sabrina and Ambrosi, Anand and Whitehead, Camden and Vitzilaios, Nikolaos},
year = {2021},
month = {04},
pages = {},
title = {Fiducial Markers for Pose Estimation: Overview, Applications and Experimental Comparison of the ARTag, AprilTag, ArUco and STag Markers},
volume = {101},
journal = {Journal of Intelligent \& Robotic Systems},
doi = {10.1007/s10846-020-01307-9}
}

@inproceedings{Deimel2014,
author = {Deimel, Raphael},
booktitle = {Robotics: Science and Systems},
pages = {1--9},
title = {{A Novel Type of Compliant , Underactuated Robotic Hand for Dexterous Grasping}},
year = {2014}
}

@article{Batsuren2019,
author = {Batsuren, Khulan and Yun, Dongwon},
doi = {10.3390/app9152967},
journal = {Applied Sciences},
number = {15},
title = {{Soft robotic gripper with chambered fingers for performing in-hand manipulation}},
volume = {9},
year = {2019}
}

@INPROCEEDINGS{Li2014,
author = {Li, Miao and Yin, Hang and Tahara, Kenji and Billard, Aude},
doi = {10.1109/ICRA.2014.6907861},
booktitle = {IEEE International Conference on Robotics and Automation},
pages = {6784--6791},
title = {{Learning object-level impedance control for robust grasping and dexterous manipulation}},
year = {2014}
}

@INPROCEEDINGS{Yuan2020,
  author={Yuan, Shenli and Shao, Lin and Yako, Connor L. and Gruebele, Alex and Salisbury, J. Kenneth},
  booktitle={IEEE/RSJ International Conference on Intelligent Robots and Systems (IROS)}, 
  title={Design and Control of Roller Grasper V2 for In-Hand Manipulation}, 
  year={2020},
  volume={},
  number={},
  pages={9151-9158},
  doi={10.1109/IROS45743.2020.9340953}}

@ARTICLE{Ijspeert2013,
  author={Ijspeert, Auke Jan and Nakanishi, Jun and Hoffmann, Heiko and Pastor, Peter and Schaal, Stefan},
  journal={Neural Computation}, 
  title={Dynamical Movement Primitives: Learning Attractor Models for Motor Behaviors}, 
  year={2013},
  volume={25},
  number={2},
  pages={328-373},
  doi={10.1162/NECO_a_00393}}

@INPROCEEDINGS{Solak2019,
  author={Solak, Gokhan and Jamone, Lorenzo},
  booktitle={IEEE/RSJ International Conference on Intelligent Robots and Systems (IROS)}, 
  title={Learning by Demonstration and Robust Control of Dexterous In-Hand Robotic Manipulation Skills}, 
  year={2019},
  volume={},
  number={},
  pages={8246-8251},
  doi={10.1109/IROS40897.2019.8967567}}

@ARTICLE{Solak2023,
  author={Solak, Gokhan and Jamone, Lorenzo},
  journal={IEEE Transactions on Haptics}, 
  title={Haptic Exploration of Unknown Objects for Robust In-Hand Manipulation}, 
  year={2023},
  volume={16},
  number={3},
  pages={400-411},
  doi={10.1109/TOH.2023.3300439}}

@ARTICLE{Yang2023,
  author={Yang, Linhan and Huang, Bidan and Li, Qingbiao and Tsai, Ya-Yen and Lee, Wang Wei and Song, Chaoyang and Pan, Jia},
  journal={IEEE Robotics and Automation Letters}, 
  title={TacGNN: Learning Tactile-Based In-Hand Manipulation With a Blind Robot Using Hierarchical Graph Neural Network}, 
  year={2023},
  volume={8},
  number={6},
  pages={3605-3612},
  doi={10.1109/LRA.2023.3264759}}

@ARTICLE{Chen2023bi,
  author={Chen, Yuanpei and Geng, Yiran and Zhong, Fangwei and Ji, Jiaming and Jiang, Jiechuang and Lu, Zongqing and Dong, Hao and Yang, Yaodong},
  journal={IEEE Transactions on Pattern Analysis and Machine Intelligence}, 
  title={Bi-DexHands: Towards Human-Level Bimanual Dexterous Manipulation}, 
  year={2023},
  volume={},
  number={},
  pages={1-15},
  doi={10.1109/TPAMI.2023.3339515}}

@article{Chen2023,
author = {Tao Chen  and Megha Tippur  and Siyang Wu  and Vikash Kumar  and Edward Adelson  and Pulkit Agrawal },
title = {Visual dexterity: In-hand reorientation of novel and complex object shapes},
journal = {Science Robotics},
volume = {8},
number = {84},
year = {2023},
doi = {10.1126/scirobotics.adc9244},
}

@Article{Ozawa2005, 
  author={R. Ozawa and S. Arimoto and S. Nakamura and Ji-Hun Bae}, 
  journal={IEEE Transactions on Robotics}, 
  title={Control of an object with parallel surfaces by a pair of finger robots without object sensing}, 
  year={2005}, 
  volume={21}, 
  number={5}, 
  pages={965-976}, 
}

@article{Singh2022,
author = {Singh, Bharat and Kumar, Rajesh and Singh, Vinay Pratap},
title = {Reinforcement Learning in Robotic Applications: A Comprehensive Survey},
year = {2022},
address = {USA},
volume = {55},
number = {2},
url = {https://doi.org/10.1007/s10462-021-09997-9},
doi = {10.1007/s10462-021-09997-9},
journal = {Artif. Intell. Rev.},
pages = {945–990},
numpages = {46},
}

@inproceedings{Sodhi2020,
title={Learning Tactile Models for Factor Graph-based Estimation},
author={Sodhi, Paloma and Kaess, Michael and Mukadam, Mustafa and Anderson, Stuart},
booktitle = {IEEE International Conference on Robotics and Automation},
year={2021}}

@INPROCEEDINGS{Koval2013,
  author={Koval, Michael C. and Dogar, Mehmet R. and Pollard, Nancy S. and Srinivasa, Siddhartha S.},
  booktitle={IEEE/RSJ Int. Conf. on Intel. Rob. and Sys.}, 
  title={Pose estimation for contact manipulation with manifold particle filters}, 
  year={2013},
  volume={},
  number={},
  pages={4541-4548},
  doi={10.1109/IROS.2013.6697009}}

@article{Falco2018,
author = {Falco, Pietro and Attawia, Abdallah and Saveriano, Matteo and Lee, Dongheui},
doi = {10.1109/lra.2018.2800110},
journal = {IEEE Robotics and Automation Letters},
number = {3},
pages = {1482--1489},
title = {{On Policy Learning Robust to Irreversible Events: An Application to Robotic In-Hand Manipulation}},
volume = {3},
year = {2018}
}

@article{Rajeswaran2017,
archivePrefix = {arXiv},
arxivId = {1709.10087},
author = {Rajeswaran, Aravind and Kumar, Vikash and Gupta, Abhishek and Vezzani, Giulia and Schulman, John and Todorov, Emanuel and Levine, Sergey},
eprint = {1709.10087},
title = {{Learning Complex Dexterous Manipulation with Deep Reinforcement Learning and Demonstrations}},
url = {http://arxiv.org/abs/1709.10087},
year = {2017}
}

@article{haldar2023teach,
  title={Teach a robot to fish: Versatile imitation from one minute of demonstrations},
  author={Haldar, Siddhant and Pari, Jyothish and Rai, Anant and Pinto, Lerrel},
  journal={arXiv preprint arXiv:2303.01497},
  year={2023}
}

@inproceedings{agarwal2023dexterous,
  title={Dexterous functional grasping},
  author={Agarwal, Ananye and Uppal, Shagun and Shaw, Kenneth and Pathak, Deepak},
  booktitle={7th Annual Conference on Robot Learning},
  year={2023}
}

@article{Kumar2016b,
archivePrefix = {arXiv},
arxivId = {1611.05095v1},
author = {Kumar, Vikash and Gupta, Abhishek and Todorov, Emanuel and Levine, Sergey},
doi = {10.1177/ToBeAssigned},
eprint = {1611.05095v1},
title = {{Learning Dexterous Manipulation Policies from Experience and Imitation}},
year = {2016}
}

@inproceedings{Gupta2016,
author = {Gupta, Abhishek and Eppner, Clemens and Levine, Sergey and Abbeel, Pieter},
booktitle = {IEEEqRSJ International Conference on Intelligent Robots and Systems},
doi = {10.1109/IROS.2016.7759557},
pages = {3786--3793},
title = {{Learning dexterous manipulation for a soft robotic hand from human demonstrations}},
year = {2016}
}

@inproceedings{Sundaralingam2018,
  title={Geometric In-Hand Regrasp Planning: Alternating Optimization of Finger Gaits and In-Grasp Manipulation},
  author={Balakumar Sundaralingam and Tucker Hermans},
  booktitle={IEEE International Conference on Robotics and Automation (ICRA)},
  year={2018},
  pages={231-238}
}

@InProceedings{Sintov2020a,
  title = 	 {Tools for Data-driven Modeling of Within-Hand Manipulation with Underactuated Adaptive Hands},
  author =       {Sintov, Avishai and Kimmel, Andrew and Wen, Bowen and Boularias, Abdeslam and Bekris, Kostas},
  booktitle = 	 {Conference on Learning for Dynamics and Control},
  pages = 	 {771--780},
  year = 	 {2020},
  editor = 	 {Bayen, Alexandre M. and Jadbabaie, Ali and Pappas, George and Parrilo, Pablo A. and Recht, Benjamin and Tomlin, Claire and Zeilinger, Melanie},
  volume = 	 {120},
  series = 	 {Proceedings of Machine Learning Research},
  month = 	 {10--11 Jun},
  publisher =    {PMLR},
}

@article{Liu2024RealDexTH,
  title={RealDex: Towards Human-like Grasping for Robotic Dexterous Hand},
  author={Yumeng Liu and Yaxun Yang and Youzhuo Wang and Xiaofei Wu and Jiamin Wang and Yichen Yao and S{\"o}ren Schwertfeger and Sibei Yang and Wenping Wang and Jingyi Yu and Xuming He and Yuexin Ma},
  journal={ArXiv},
  year={2024},
  volume={abs/2402.13853},
}

@article{khazatsky2024,
      title={{DROID}: A Large-Scale In-The-Wild Robot Manipulation Dataset}, 
      author={Alexander Khazatsky and Karl Pertsch and Suraj Nair and Ashwin Balakrishna and Sudeep Dasari and Siddharth Karamcheti and Soroush Nasiriany and Mohan Kumar Srirama and Lawrence Yunliang Chen and Kirsty Ellis and Peter David Fagan and Joey Hejna and Masha Itkina and Marion Lepert and Yecheng Jason Ma and Patrick Tree Miller and Jimmy Wu and Suneel Belkhale and Shivin Dass and Huy Ha and Arhan Jain and Abraham Lee and Youngwoon Lee and Marius Memmel and Sungjae Park and Ilija Radosavovic and Kaiyuan Wang and Albert Zhan and Kevin Black and Cheng Chi and Kyle Beltran Hatch and Shan Lin and Jingpei Lu and Jean Mercat and Abdul Rehman and Pannag R Sanketi and Archit Sharma and Cody Simpson and Quan Vuong and Homer Rich Walke and Blake Wulfe and Ted Xiao and Jonathan Heewon Yang and Arefeh Yavary and Tony Z. Zhao and Christopher Agia and Rohan Baijal and Mateo Guaman Castro and Daphne Chen and Qiuyu Chen and Trinity Chung and Jaimyn Drake and Ethan Paul Foster and Jensen Gao and David Antonio Herrera and Minho Heo and Kyle Hsu and Jiaheng Hu and Donovon Jackson and Charlotte Le and Yunshuang Li and Kevin Lin and Roy Lin and Zehan Ma and Abhiram Maddukuri and Suvir Mirchandani and Daniel Morton and Tony Nguyen and Abigail O'Neill and Rosario Scalise and Derick Seale and Victor Son and Stephen Tian and Emi Tran and Andrew E. Wang and Yilin Wu and Annie Xie and Jingyun Yang and Patrick Yin and Yunchu Zhang and Osbert Bastani and Glen Berseth and Jeannette Bohg and Ken Goldberg and Abhinav Gupta and Abhishek Gupta and Dinesh Jayaraman and Joseph J Lim and Jitendra Malik and Roberto Martín-Martín and Subramanian Ramamoorthy and Dorsa Sadigh and Shuran Song and Jiajun Wu and Michael C. Yip and Yuke Zhu and Thomas Kollar and Sergey Levine and Chelsea Finn},
      year={2024},
      eprint={2403.12945},
      journal={arXiv},
      primaryClass={cs.RO},
      url={https://arxiv.org/abs/2403.12945}, 
}

@article{Shi2017,
author = {Shi, Jian and Woodruff, J. Zachary and Umbanhowar, Paul B. and Lynch, Kevin M.},
doi = {10.1109/TRO.2017.2693391},
journal = {IEEE Transactions on Robotics},
number = {4},
pages = {778--795},
title = {{Dynamic In-Hand Sliding Manipulation}},
volume = {33},
year = {2017}
}

@inproceedings{liarokapis2016post,
  title={Post-contact, in-hand object motion compensation for compliant and underactuated hands},
  author={Liarokapis, Minas and Dollar, Aaron M},
  booktitle={IEEE International Symposium on Robot and Human Interactive Communication (RO-MAN)},
  pages={986--993},
  year={2016}
}

@article{Fan2017,
author = {Fan, Yongxiang and Gao, Wei and Chen, Wenjie and Tomizuka, Masayoshi},
doi = {10.1016/j.ifacol.2017.08.1831},
issn = {24058963},
journal = {IFAC-PapersOnLine},
number = {1},
pages = {12765--12772},
publisher = {Elsevier B.V.},
title = {{Real-Time Finger Gaits Planning for Dexterous Manipulation}},
url = {https://doi.org/10.1016/j.ifacol.2017.08.1831},
volume = {50},
year = {2017}
}

@article{Sintov2019,
author = {Sintov, Avishai and Morgan, Andrew S. and Kimmel, Andrew and Dollar, Aaron M. and Bekris, Kostas E. and Boularias, Abdeslam},
doi = {10.1109/LRA.2019.2894875},
journal = {IEEE Robotics and Automation Letters},
number = {2},
pages = {1287--1294},
title = {{Learning a State Transition Model of an Underactuated Adaptive Hand}},
volume = {4},
year = {2019}
}

@inproceedings{Nematollahi2022,
    author  = {Iman Nematollahi and Erick Rosete-Beas and Seyed Mahdi B. Azad and Raghu Rajan and Frank Hutter and Wolfram Burgard},
    title   = {T3VIP: Transformation-based 3D Video Prediction},
    booktitle = {IEEE/RSJ International Conference on Intelligent Robots and Systems (IROS)},
    year = {2022},
}

@inproceedings{wettels2009multi,
  title={Multi-modal synergistic tactile sensing},
  author={Wettels, Nicholas and Fishel, Jeremy A and Su, Z and Lin, Chia Hsien and Loeb, Gerald E and SynTouch, LLC},
  booktitle={Tactile sensing in humanoids—Tactile sensors and beyond workshop, 9th IEEE-RAS international conference on humanoid robots},
  year={2009}
}

@article{kroemer2021,
  title={A review of robot learning for manipulation: Challenges, representations, and algorithms},
  author={Kroemer, Oliver and Niekum, Scott and Konidaris, George},
  journal={The Journal of Machine Learning Research},
  volume={22},
  number={1},
  pages={1395--1476},
  year={2021},
}

@inproceedings{cheng2009novel,
  title={A novel highly-twistable tactile sensing array using extendable spiral electrodes},
  author={Cheng, M-Y and Tsao, C-M and Lai, Y-T and Yang, Y-J},
  booktitle={IEEE International Conference on Micro Electro Mechanical Systems},
  pages={92--95},
  year={2009},
}

@article{Huang2021,
      title={Generalization in Dexterous Manipulation via Geometry-Aware Multi-Task Learning}, 
      author={Wenlong Huang and Igor Mordatch and Pieter Abbeel and Deepak Pathak},
      year={2021},
      eprint={2111.03062},
      journal={arXiv},
}

@INPROCEEDINGS{Gupta2021,
  author={Gupta, Abhishek and Yu, Justin and Zhao, Tony Z. and Kumar, Vikash and Rovinsky, Aaron and Xu, Kelvin and Devlin, Thomas and Levine, Sergey},
  booktitle={IEEE International Conference on Robotics and Automation (ICRA)}, 
  title={Reset-Free Reinforcement Learning via Multi-Task Learning: Learning Dexterous Manipulation Behaviors without Human Intervention}, 
  year={2021},
  volume={},
  number={},
  pages={6664-6671},
  doi={10.1109/ICRA48506.2021.9561384}}

@article{yousef2011tactile,
  title={Tactile sensing for dexterous in-hand manipulation in robotics—A review},
  author={Yousef, Hanna and Boukallel, Mehdi and Althoefer, Kaspar},
  journal={Sensors and Actuators A: physical},
  volume={167},
  number={2},
  pages={171--187},
  year={2011},
  publisher={Elsevier}
}

@article{Chavan-Dafle2020,
archivePrefix = {arXiv},
arxivId = {arXiv:1810.00219v1},
author = {Chavan-Dafle, Nikhil and Holladay, Rachel and Rodriguez, Alberto},
doi = {10.1177/0278364919880257},
eprint = {arXiv:1810.00219v1},
journal = {International Journal of Robotics Research},
number = {2-3},
pages = {163--182},
title = {{Planar in-hand manipulation via motion cones}},
volume = {39},
year = {2020}
}

@INPROCEEDINGS{Li2020,
  author={Li, Tingguang and Srinivasan, Krishnan and Meng, Max Qing-Hu and Yuan, Wenzhen and Bohg, Jeannette},
  booktitle={IEEE International Conference on Robotics and Automation (ICRA)}, 
  title={Learning Hierarchical Control for Robust In-Hand Manipulation}, 
  year={2020},
  volume={},
  number={},
  pages={8855-8862},
  doi={10.1109/ICRA40945.2020.9197343}}

@ARTICLE{Liu2020,
  author={Liu, Huan and Zhao, Longhai and Siciliano, Bruno and Ficuciello, Fanny},
  journal={IEEE Robotics and Automation Letters}, 
  title={Modeling, Optimization, and Experimentation of the ParaGripper for In-Hand Manipulation Without Parasitic Rotation}, 
  year={2020},
  volume={5},
  number={2},
  pages={3011-3018},
  doi={10.1109/LRA.2020.2974419}}

@INPROCEEDINGS{Calli2018, 
author={B. Calli and K. Srinivasan and A. Morgan and A. M. Dollar}, 
booktitle={IEEE International Conference on Robotics and Automation (ICRA)}, 
title={Learning Modes of Within-Hand Manipulation}, 
year={2018}, 
volume={}, 
number={}, 
pages={3145-3151}, 
doi={10.1109/ICRA.2018.8461187}, 
ISSN={2577-087X}, 
}

@INPROCEEDINGS{Calli2017, 
  author={B. Calli and A. M. Dollar}, 
  booktitle={IEEE International Conference on Robotics and Automation}, 
  title={Vision-based model predictive control for within-hand precision manipulation with underactuated grippers}, 
  year={2017}, 
  volume={}, 
  number={}, 
  pages={2839-2845}, 
}

@INPROCEEDINGS{Tao2023,
  author={Tao, Lingfeng and Zhang, Jiucai and Bowman, Michael and Zhang, Xiaoli},
  booktitle={IEEE International Conference on Robotics and Automation (ICRA)}, 
  title={A Multi-Agent Approach for Adaptive Finger Cooperation in Learning-based In-Hand Manipulation}, 
  year={2023},
  volume={},
  number={},
  pages={3897-3903},
  doi={10.1109/ICRA48891.2023.10160909}}

@inproceedings{nagabandi2020deep,
  title={Deep dynamics models for learning dexterous manipulation},
  author={Nagabandi, Anusha and Konolige, Kurt and Levine, Sergey and Kumar, Vikash},
  booktitle={Conference on Robot Learning},
  pages={1101--1112},
  year={2020},
  organization={PMLR}
}

@ARTICLE{Pourpanah2023,
  author={Pourpanah, Farhad and Abdar, Moloud and Luo, Yuxuan and Zhou, Xinlei and Wang, Ran and Lim, Chee Peng and Wang, Xi-Zhao and Wu, Q. M. Jonathan},
  journal={IEEE Transactions on Pattern Analysis and Machine Intelligence}, 
  title={A Review of Generalized Zero-Shot Learning Methods}, 
  year={2023},
  volume={45},
  number={4},
  pages={4051-4070},
  doi={10.1109/TPAMI.2022.3191696}}

@article{VanHoof2015,
author = {{Van Hoof}, Herke and Hermans, Tucker and Neumann, Gerhard and Peters, Jan},
doi = {10.1109/HUMANOIDS.2015.7363524},
issn = {21640580},
journal = {IEEE-RAS International Conference on Humanoid Robots},
pages = {121--127},
title = {{Learning robot in-hand manipulation with tactile features}},
volume = {2015-Decem},
year = {2015}
}

@INPROCEEDINGS{M.Okamura2000,
  author={Okamura, A.M. and Smaby, N. and Cutkosky, M.R.},
  booktitle={IEEE International Conference on Robotics and Automation. }, 
  title={An overview of dexterous manipulation}, 
  year={2000},
  volume={1},
  number={},
  pages={255-262}}

@INPROCEEDINGS{Ma2011,
  author={Ma, Raymond R. and Dollar, Aaron M.},
  booktitle={International Conference on Advanced Robotics (ICAR)}, 
  title={On dexterity and dexterous manipulation}, 
  year={2011},
  volume={},
  number={},
  pages={1-7}}

@INPROCEEDINGS{Butterfass2001,
  author={Butterfass, J. and Grebenstein, M. and Liu, H. and Hirzinger, G.},
  booktitle={IEEE International Conference on Robotics and Automation}, 
  title={DLR-Hand II: next generation of a dexterous robot hand}, 
  year={2001},
  volume={1},
  number={},
  pages={109-114 vol.1},
  doi={10.1109/ROBOT.2001.932538}}

@INPROCEEDINGS{Schulz2001,
  author={Schulz, S. and Pylatiuk, C. and Bretthauer, G.},
  booktitle={IEEE International Conference on Robotics and Automation}, 
  title={A new ultralight anthropomorphic hand}, 
  year={2001},
  volume={3},
  number={},
  pages={2437-2441 vol.3},
  doi={10.1109/ROBOT.2001.932988}}

@INPROCEEDINGS{Lovchik1999,
  author={Lovchik, C.S. and Diftler, M.A.},
  booktitle={IEEE International Conference on Robotics and Automation}, 
  title={The Robonaut hand: a dexterous robot hand for space}, 
  year={1999},
  volume={2},
  number={},
  pages={907-912 vol.2},
  doi={10.1109/ROBOT.1999.772420}
}

@inproceedings{Mouri2002,
  title={Anthropomorphic Robot Hand : Gifu Hand III},
  author={Tetsuya Mouri},
  year={2002},
  booktitle={International Conf. Control, Automation and Systems},
  pages={1288-1293}
}

@INPROCEEDINGS{Lotti2004,
  author={Lotti, F. and Tiezzi, P. and Vassura, G. and Biagiotti, L. and Melchiorri, C.},
  booktitle={IEEE International Conference on Robotics and Automation}, 
  title={UBH 3: an anthropomorphic hand with simplified endo-skeletal structure and soft continuous fingerpads}, 
  year={2004},
  volume={5},
  number={},
  pages={4736-4741},
  doi={10.1109/ROBOT.2004.1302466}}

@article{Delgado2017,
title = {In-hand recognition and manipulation of elastic objects using a servo-tactile control strategy},
journal = {Robotics and Computer-Integrated Manufacturing},
volume = {48},
pages = {102-112},
year = {2017},
doi = {https://doi.org/10.1016/j.rcim.2017.03.002},
url = {https://www.sciencedirect.com/science/article/pii/S0736584515300995},
author = {A. Delgado and C.A. Jara and F. Torres}
}

@article{Sun2022,
author = {Sun, Huanbo and Kuchenbecker, Katherine and Martius, Georg},
year = {2022},
pages = {135–145},
volume = {4},
journal = {Nature Machine Intelligence},
title = {A soft thumb-sized vision-based sensor with accurate all-round force perception}
}

@ARTICLE{Bimbo2016,
  author={Bimbo, Joao and Luo, Shan and Althoefer, Kaspar and Liu, Hongbin},
  journal={IEEE Robotics and Automation Letters}, 
  title={In-Hand Object Pose Estimation Using Covariance-Based Tactile To Geometry Matching},
  year={2016},
  volume={1},
  number={1},
  pages={570-577},
  doi={10.1109/LRA.2016.2517244}}

@inproceedings{Carter2005,
author = {Carter, Jim and Fourney, David},
year = {2005},
month = {01},
pages = {84-92},
title = {Research based tactile and haptic interaction guidelines},
booktitle={Proceedings of Guidelines On Tactile and Haptic Interactions},  
}

@article{Jain2019,
author = {Jain, Divye and Li, Andrew and Singhal, Shivam and Rajeswaran, Aravind and Kumar, Vikash and Todorov, Emanuel},
doi = {10.1109/icra.2019.8794033},
journal = {IEEE International Conference on Robotics and Automation (ICRA)},
pages = {3636--3643},
title = {{Learning Deep Visuomotor Policies for Dexterous Hand Manipulation}},
year = {2019}
}

@InProceedings{Choi2017,
author="Choi, Changhyun
and Del Preto, Joseph
and Rus, Daniela",
editor="Kuli{\'{c}}, Dana
and Nakamura, Yoshihiko
and Khatib, Oussama
and Venture, Gentiane",
title="Using Vision for Pre- and Post-grasping Object Localization for Soft Hands",
booktitle="International Symposium on Experimental Robotics",
year="2017",
address="Cham",
pages="601--612",
}

@article{OpenAI2019,
archivePrefix = {arXiv},
arxivId = {1910.07113},
author = {OpenAI and Akkaya, Ilge and Andrychowicz, Marcin and Chociej, Maciek and Litwin, Mateusz and McGrew, Bob and Petron, Arthur and Paino, Alex and Plappert, Matthias and Powell, Glenn and Ribas, Raphael and Schneider, Jonas and Tezak, Nikolas and Tworek, Jerry and Welinder, Peter and Weng, Lilian and Yuan, Qiming and Zaremba, Wojciech and Zhang, Lei},
eprint = {1910.07113},
pages = {1--51},
title = {{Solving Rubik's Cube with a Robot Hand}},
url = {http://arxiv.org/abs/1910.07113},
year = {2019}
}

@article{Andrychowicz2017,
archivePrefix = {arXiv},
arxivId = {1707.01495},
author = {Andrychowicz, Marcin and Wolski, Filip and Ray, Alex and Schneider, Jonas and Fong, Rachel and Welinder, Peter and McGrew, Bob and Tobin, Josh and Abbeel, Pieter and Zaremba, Wojciech},
eprint = {1707.01495},
number = {Nips},
title = {{Hindsight Experience Replay (HER)}},
url = {http://arxiv.org/abs/1707.01495},
year = {2017}
}

@InProceedings{qi2022,
   author={Qi, Haozhi and Kumar, Ashish and Calandra, Roberto and Ma, Yi and Malik, Jitendra},
   title={{In-Hand Object Rotation via Rapid Motor Adaptation}},
   booktitle={Conference on Robot Learning (CoRL)},
   year={2022}
  }

@InProceedings{Qi2023,
  title = 	 {General In-hand Object Rotation with Vision and Touch},
  author =       {Qi, Haozhi and Yi, Brent and Suresh, Sudharshan and Lambeta, Mike and Ma, Yi and Calandra, Roberto and Malik, Jitendra},
  booktitle = 	 {Conference on Robot Learning},
  pages = 	 {2549--2564},
  year = 	 {2023},
  editor = 	 {Tan, Jie and Toussaint, Marc and Darvish, Kourosh},
  volume = 	 {229},
}

@article{luo2023,
      title={Progressive Transfer Learning for Dexterous In-Hand Manipulation with Multi-Fingered Anthropomorphic Hand}, 
      author={Yongkang Luo and Wanyi Li and Peng Wang and Haonan Duan and Wei Wei and Jia Sun},
      year={2023},
      archivePrefix={arXiv preprint arXiv:2304.09526},
      primaryClass={cs.RO}
}

@InProceedings{Yin2023,
      title          = {Rotating without Seeing: Towards In-hand Dexterity through Touch },
      author         = {Yin, Zhao-Heng and Huang, Binghao and Qin, Yuzhe and Chen, Qifeng and Wang, Xiaolong},
      booktitle        = {Robotics: Science and Systems},
      year           = {2023},
    }

@inproceedings{taylor2022gelslim,
  title={GelSlim 3.0: High-resolution measurement of shape, force and slip in a compact tactile-sensing finger},
  author={Taylor, Ian H and Dong, Siyuan and Rodriguez, Alberto},
  booktitle={IEEE International Conference on Robotics and Automation (ICRA)},
  pages={10781--10787},
  year={2022},
}

@INPROCEEDINGS{He2015,
  author={He, Junhu and Pu, Sicong and Zhang, Jianwei},
  booktitle={IEEE International Conference on Robotics and Biomimetics (ROBIO)}, 
  title={Haptic and visual perception in in-hand manipulation system}, 
  year={2015},
  volume={},
  number={},
  pages={303-308},
  doi={10.1109/ROBIO.2015.7418784}}

@article{Funabashi2019,
author = {Funabashi, Satoshi and Schmitz, Alexander and Sato, Takashi and Somlor, Sophon and Sugano, Shigeki},
doi = {10.1109/HUMANOIDS.2018.8624961},
journal = {IEEE-RAS International Conference on Humanoid Robots},
pages = {768--775},
title = {{Versatile In-Hand Manipulation of Objects with Different Sizes and Shapes Using Neural Networks}},
volume = {2018-Novem},
year = {2019}
}

@inproceedings{Nguyen2019,
author = {Nguyen, Hai and La, Hung Manh},
doi = {10.1109/IRC.2019.00120},
booktitle={IEEE International Conference on Robotic Computing (IRC)},
pages = {590--595},
title = {{Review of Deep Reinforcement Learning for Robot Manipulation}},
year = {2019}
}

@INPROCEEDINGS{Handa2023,
  author={Handa, Ankur and Allshire, Arthur and Makoviychuk, Viktor and Petrenko, Aleksei and Singh, Ritvik and Liu, Jingzhou and Makoviichuk, Denys and Van Wyk, Karl and Zhurkevich, Alexander and Sundaralingam, Balakumar and Narang, Yashraj},
  booktitle={IEEE International Conference on Robotics and Automation (ICRA)}, 
  title={DeXtreme: Transfer of Agile In-hand Manipulation from Simulation to Reality}, 
  year={2023},
  volume={},
  number={},
  pages={5977-5984},
  doi={10.1109/ICRA48891.2023.10160216}}

@article{RoblesDeLaTorre2001,
  title={Force can overcome object geometry in the perception of shape through active touch},
  author={Gabriel Robles-De-La-Torre and Vincent Hayward},
  journal={Nature},
  year={2001},
  volume={412},
  pages={445-448}
}

@article{okada1979object,
  title={Object-handling system for manual industry},
  author={Okada, Tokuji},
  journal={IEEE Transactions on Systems, Man, and Cybernetics},
  volume={9},
  number={2},
  pages={79--89},
  year={1979},
  publisher={IEEE}
}

@article{Townsend2000,
author = {Townsend, William},
year = {2000},
month = {06},
pages = {181-188},
title = {The BarrettHand grasper – programmably flexible part handling and assembly},
volume = {27},
journal = {Industrial Robot: An International Journal},
doi = {10.1108/01439910010371597}
}

@INPROCEEDINGS{Jacobsen1986,
  author={Jacobsen, S. and Iversen, E. and Knutti, D. and Johnson, R. and Biggers, K.},
  booktitle={IEEE International Conference on Robotics and Automation}, 
  title={Design of the Utah/M.I.T. Dextrous Hand}, 
  year={1986},
  volume={3},
  number={},
  pages={1520-1532},
  doi={10.1109/ROBOT.1986.1087395}}

@InProceedings{Toskov2023,
  title = 	 {In-Hand Gravitational Pivoting Using Tactile Sensing},
  author =       {Toskov, Jason and Newbury, Rhys and Mukadam, Mustafa and Kulic, Dana and Cosgun, Akansel},
  booktitle = 	 {Proceedings of The 6th Conference on Robot Learning},
  pages = 	 {2284--2293},
  year = 	 {2023},
  editor = 	 {Liu, Karen and Kulic, Dana and Ichnowski, Jeff},
  volume = 	 {205},
  series = 	 {Proceedings of Machine Learning Research},
  publisher =    {PMLR},
  url = 	 {https://proceedings.mlr.press/v205/toskov23a.html},
}

@article{Dimou2023,
  title={Robotic hand synergies for in-hand regrasping driven by object information},
  author={Dimitrios Dimou and Jos{\'e} Santos-Victor and Plinio Moreno},
  journal={Autonomous Robots},
  year={2023},
  volume={47},
  pages={453 - 464},
  url={https://api.semanticscholar.org/CorpusID:258093490}
}

@ARTICLE{Calli2015, 
author={B. {Calli} and A. {Walsman} and A. {Singh} and S. {Srinivasa} and P. {Abbeel} and A. M. {Dollar}}, 
journal={IEEE Robotics Automation Magazine}, 
title={Benchmarking in Manipulation Research: Using the Yale-CMU-Berkeley Object and Model Set}, 
year={2015}, 
volume={22}, 
number={3}, 
pages={36-52}}

@INPROCEEDINGS{Dafle2014,
  author={Dafle, Nikhil Chavan and Rodriguez, Alberto and Paolini, Robert and Tang, Bowei and Srinivasa, Siddhartha S. and Erdmann, Michael and Mason, Matthew T. and Lundberg, Ivan and Staab, Harald and Fuhlbrigge, Thomas},
  booktitle={IEEE International Conference on Robotics and Automation (ICRA)}, 
  title={Extrinsic dexterity: In-hand manipulation with external forces}, 
  year={2014},
  volume={},
  number={},
  pages={1578-1585}}

@book{Mason1985,
author = {Mason, Matthew T. and Salisbury, J. Kenneth},
title = {Robot Hands and the Mechanics of Manipulation},
year = {1985},
publisher = {MIT Press},
address = {Cambridge, MA, USA}
}

@INPROCEEDINGS{Guo2017,
  author={Guo, Menglong and Gealy, David V. and Liang, Jacky and Mahler, Jeffrey and Goncalves, Aimee and McKinley, Stephen and Ojea, Juan Aparicio and Goldberg, Ken},
  booktitle={IEEE International Conference on Robotics and Automation (ICRA)}, 
  title={Design of parallel-jaw gripper tip surfaces for robust grasping}, 
  year={2017},
  volume={},
  number={},
  pages={2831-2838}}

@INPROCEEDINGS{Tournassoud1987,
  author={Tournassoud, P. and Lozano-Perez, T. and Mazer, E.},
  booktitle={IEEE International Conference on Robotics and Automation}, 
  title={Regrasping}, 
  year={1987},
  volume={4},
  number={},
  pages={1924-1928},
  doi={10.1109/ROBOT.1987.1087910}}

@INPROCEEDINGS{Sintov2016,
  author={Sintov, Avishai and Shapiro, Amir},
  booktitle={IEEE International Conference on Robotics and Automation (ICRA)}, 
  title={Swing-up regrasping algorithm using energy control}, 
  year={2016},
  volume={},
  number={},
  pages={4888-4893}}

@article{shi2017dynamic,
  title={Dynamic in-hand sliding manipulation},
  author={Shi, Jian and Woodruff, J Zachary and Umbanhowar, Paul B and Lynch, Kevin M},
  journal={IEEE Transactions on Robotics},
  volume={33},
  number={4},
  pages={778--795},
  year={2017},
  publisher={IEEE}
}

@ARTICLE{Shi2020,
  author={Shi, Jian and Weng, Huan and Lynch, Kevin M.},
  journal={IEEE Access}, 
  title={In-Hand Sliding Regrasp With Spring-Sliding Compliance and External Constraints}, 
  year={2020},
  volume={8},
  number={},
  pages={88729-88744},
  doi={10.1109/ACCESS.2020.2991382}}

@article{Rus1999,
author = {Daniela Rus},
title ={In-Hand Dexterous Manipulation of Piecewise-Smooth 3D Objects},
journal = {The International Journal of Robotics Research},
volume = {18},
number = {4},
pages = {355-381},
year = {1999}
}

@INPROCEEDINGS{Xu2007,
  author={Xu, Jijie and Koo, T. John and Li, Zexiang},
  booktitle={IEEE/RSJ International Conference on Intelligent Robots and Systems}, 
  title={Finger gaits planning for multifingered manipulation}, 
  year={2007},
  volume={},
  number={},
  pages={2932-2937},
  doi={10.1109/IROS.2007.4399189}}

@INPROCEEDINGS{Lozano-Perez1987,
  author={Lozano-Perez, T. and Jones, J. and Mazer, E. and O'Donnell, P. and Grimson, W. and Tournassoud, P. and Lanusse, A.},
  booktitle={IEEE International Conference on Robotics and Automation}, 
  title={Handey: A robot system that recognizes, plans, and manipulates}, 
  year={1987},
  volume={4},
  number={},
  pages={843-849},
  doi={10.1109/ROBOT.1987.1087847}}

@INPROCEEDINGS{Furukawa2006,
  author={Furukawa, N. and Namiki, A. and Taku, S. and Ishikawa, M.},
  booktitle={IEEE International Conference on Robotics and Automation}, 
  title={Dynamic regrasping using a high-speed multifingered hand and a high-speed vision system}, 
  year={2006},
  volume={},
  number={},
  pages={181-187},
  doi={10.1109/ROBOT.2006.1641181}}

@inproceedings{Kokic2019,
    author = {Kokic, Mia and Kragic, Danica and Bohg, Jeannette},
    year = {2019},
    booktitle={IEEE/RSJ International Conference on Intelligent Robots and Systems (IROS)}, 
    month = {11},
    pages = {3980-3987},
    title = {Learning to Estimate Pose and Shape of Hand-Held Objects from RGB Images},
    doi = {10.1109/IROS40897.2019.8967961}
}

@INPROCEEDINGS{Han1997,
  author={Han, L. and Guan, Y.S. and Li, Z.X. and Shi, Q. and Trinkle, J.C.},
  booktitle={IEEE International Conference on Robotics and Automation}, 
  title={Dextrous manipulation with rolling contacts}, 
  year={1997},
  volume={2},
  number={},
  pages={992-997 vol.2}}

@Article{Nahum2022,
 title = {Robotic manipulation of thin objects within off-the-shelf parallel grippers with a vibration finger},
journal = {Mechanism and Machine Theory},
volume = {177},
pages = {105032},
year = {2022},
author = {Noam Nahum and Avishai Sintov},
}

@article{mnih2015human,
  title={Human-level control through deep reinforcement learning},
  author={Mnih, Volodymyr and Kavukcuoglu, Koray and Silver, David and Rusu, Andrei A and Veness, Joel and Bellemare, Marc G and Graves, Alex and Riedmiller, Martin and Fidjeland, Andreas K and Ostrovski, Georg and others},
  journal={Nature},
  volume={518},
  number={7540},
  pages={529},
  year={2015},
  publisher={Nature Publishing Group}
}

@article{watkins1992q,
  title={Q-learning},
  author={Watkins, Christopher JCH and Dayan, Peter},
  journal={Machine learning},
  volume={8},
  number={3-4},
  pages={279--292},
  year={1992},
  publisher={Springer}
}

@article{mnih2013playing,
  title={Playing atari with deep reinforcement learning},
  author={Mnih, Volodymyr and Kavukcuoglu, Koray and Silver, David and Graves, Alex and Antonoglou, Ioannis and Wierstra, Daan and Riedmiller, Martin},
  journal={arXiv preprint arXiv:1312.5602},
  year={2013}
}

@article{lillicrap2015continuous,
  title={Continuous control with deep reinforcement learning},
  author={Lillicrap, Timothy P and Hunt, Jonathan J and Pritzel, Alexander and Heess, Nicolas and Erez, Tom and Tassa, Yuval and Silver, David and Wierstra, Daan},
  journal={International Conference on Learning Representations (ICLR)},
  year={2016}
}

@article{Ma2017YaleOP,
  title={Yale OpenHand Project: Optimizing Open-Source Hand Designs for Ease of Fabrication and Adoption},
  author={Raymond R. Ma and Aaron M. Dollar},
  journal={IEEE Robotics and Autumation Magezine},
  year={2017},
  volume={24},
  pages={32-40}
}

@inproceedings{Odhner2011,
author = {Odhner, Lael U. and Dollar, Aaron M.},
booktitle={IEEE Int. Conf. on Rob. and Aut.}, 
pages = {5254--5260},
title = {{Dexterous manipulation with underactuated elastic hands}},
year = {2011}
}

@article{Dollar2010,
author = {Aaron M. Dollar and Robert D. Howe},
title ={The Highly Adaptive SDM Hand: Design and Performance Evaluation},
journal = {The International Journal of Robotics Research},
volume = {29},
number = {5},
pages = {585-597},
year = {2010},
}

@article{antonova2017reinforcement,
  title={Reinforcement learning for pivoting task},
  author={Antonova, Rika and Cruciani, Silvia and Smith, Christian and Kragic, Danica},
  journal={arXiv preprint arXiv:1703.00472},
  year={2017}
}

@article{levine2016end,
  title={End-to-end training of deep visuomotor policies},
  author={Levine, Sergey and Finn, Chelsea and Darrell, Trevor and Abbeel, Pieter},
  journal={The Journal of Machine Learning Research},
  volume={17},
  number={1},
  pages={1334--1373},
  year={2016},
  publisher={JMLR. org}
}

@inproceedings{todorov2012mujoco,
  title={Mujoco: A physics engine for model-based control},
  author={Todorov, Emanuel and Erez, Tom and Tassa, Yuval},
  booktitle={2012 IEEE/RSJ International Conference on Intelligent Robots and Systems},
  pages={5026--5033},
  year={2012},
  organization={IEEE}
}

@article{Gaspar2018,
title = {Skill learning and action recognition by arc-length dynamic movement primitives},
journal = {Robotics and Autonomous Systems},
volume = {100},
pages = {225-235},
year = {2018},
issn = {0921-8890},
doi = {https://doi.org/10.1016/j.robot.2017.11.012},
url = {https://www.sciencedirect.com/science/article/pii/S0921889017302695},
author = {Timotej Gašpar and Bojan Nemec and Jun Morimoto and Aleš Ude},
}

@inproceedings{Duan2017,
 author = {Duan, Yan and Andrychowicz, Marcin and Stadie, Bradly and Jonathan Ho, OpenAI and Schneider, Jonas and Sutskever, Ilya and Abbeel, Pieter and Zaremba, Wojciech},
 booktitle = {Advances in Neural Information Processing Systems},
 editor = {I. Guyon and U. Von Luxburg and S. Bengio and H. Wallach and R. Fergus and S. Vishwanathan and R. Garnett},
 pages = {},
 publisher = {Curran Associates, Inc.},
 title = {One-Shot Imitation Learning},
 url = {https://proceedings.neurips.cc/paper_files/paper/2017/file/ba3866600c3540f67c1e9575e213be0a-Paper.pdf},
 volume = {30},
 year = {2017}
}

@INPROCEEDINGS{Vina2015,
  author={Viña B., Francisco E. and Karayiannidis, Yiannis and Pauwels, Karl and Smith, Christian and Kragic, Danica},
  booktitle={IEEE/RSJ International Conference on Intelligent Robots and Systems (IROS)}, 
  title={In-hand manipulation using gravity and controlled slip}, 
  year={2015},
  volume={},
  number={},
  pages={5636-5641}}

@INPROCEEDINGS{Cruciani2017,
  author={Cruciani, Silvia and Smith, Christian},
  booktitle={IEEE/RSJ International Conference on Intelligent Robots and Systems (IROS)}, 
  title={In-hand manipulation using three-stages open loop pivoting}, 
  year={2017},
  volume={},
  number={},
  pages={1244-1251},
  doi={10.1109/IROS.2017.8202299}}

@misc{rapisarda2019reinforcement,
  title={Reinforcement Learning for Dexterity Transfer Between Manipulators},
  author={Rapisarda, Carlo},
  year={2019}
}

@article{hussein2017imitation,
  title={Imitation learning: A survey of learning methods},
  author={Hussein, Ahmed and Gaber, Mohamed Medhat and Elyan, Eyad and Jayne, Chrisina},
  journal={ACM Computing Surveys (CSUR)},
  volume={50},
  number={2},
  pages={1--35},
  year={2017},
  publisher={ACM New York, NY, USA}
}

@article{andrychowicz2020learning,
  title={Learning dexterous in-hand manipulation},
  author={Andrychowicz, OpenAI: Marcin and Baker, Bowen and Chociej, Maciek and Jozefowicz, Rafal and McGrew, Bob and Pachocki, Jakub and Petron, Arthur and Plappert, Matthias and Powell, Glenn and Ray, Alex and others},
  journal={The International Journal of Robotics Research},
  volume={39},
  number={1},
  pages={3--20},
  year={2020},
  publisher={SAGE Publications Sage UK: London, England}
}

@inproceedings{chen2021simple,
  title={A Simple Method for Complex In-hand Manipulation},
  author={Chen, Tao and Xu, Jie and Agrawal, Pulkit},
  booktitle={Annual Conference on Robot Learning},
  volume={3},
  year={2021}
}

@INPROCEEDINGS{Wen2020,
  author={Wen, Bowen and Mitash, Chaitanya and Soorian, Sruthi and Kimmel, Andrew and Sintov, Avishai and Bekris, Kostas E.},
  booktitle={IEEE International Conference on Robotics and Automation (ICRA)}, 
  title={Robust, Occlusion-aware Pose Estimation for Objects Grasped by Adaptive Hands}, 
  year={2020},
  volume={},
  number={},
  pages={6210-6217},
  doi={10.1109/ICRA40945.2020.9197350}}

@ARTICLE{Pfanne2020,
  author={Pfanne, Martin and Chalon, Maxime and Stulp, Freek and Ritter, Helge and Albu-Schäffer, Alin},
  journal={IEEE Robotics and Automation Letters}, 
  title={Object-Level Impedance Control for Dexterous In-Hand Manipulation}, 
  year={2020},
  volume={5},
  number={2},
  pages={2987-2994},
  doi={10.1109/LRA.2020.2974702}}

@ARTICLE{Pagoli2021,
  author={Pagoli, Amir and Chapelle, Frédéric and Corrales, Juan Antonio and Mezouar, Youcef and Lapusta, Yuri},
  journal={IEEE Robotics and Automation Letters}, 
  title={A Soft Robotic Gripper With an Active Palm and Reconfigurable Fingers for Fully Dexterous In-Hand Manipulation *}, 
  year={2021},
  volume={6},
  number={4},
  pages={7706-7713},
  doi={10.1109/LRA.2021.3098803}}

@inproceedings{chen2022,
title={A System for General In-Hand Object Re-Orientation},
author={Tao Chen and Jie Xu and Pulkit Agrawal},
booktitle={Conference on Robot Learning },
year={2022},
url={https://openreview.net/forum?id=7uSBJDoP7tY}
}

@INPROCEEDINGS{Khandate2023, 
    AUTHOR    = {Gagan Khandate AND Siqi Shang AND Eric T Chang AND Tristan L Saidi AND Johnson Adams AND Matei Ciocarlie}, 
    TITLE     = {{Sampling-based Exploration for Reinforcement Learning of Dexterous Manipulation}}, 
    BOOKTITLE = {Robotics: Science and Systems}, 
    YEAR      = {2023}, 
    ADDRESS   = {Daegu, Republic of Korea}, 
}

@INPROCEEDINGS{Khandate2022,
  author={Khandate, Gagan and Haas-Heger, Maximilian and Ciocarlie, Matei},
  booktitle={IEEE International Conference on Robotics and Automation (ICRA)}, 
  title={On the Feasibility of Learning Finger-gaiting In-hand Manipulation with Intrinsic Sensing}, 
  year={2022},
  volume={},
  number={},
  pages={2752-2758},
  doi={10.1109/ICRA46639.2022.9812212}}

@InProceedings{Toledo2021,
author="Toledo, Leonardo V. O.
and Giardini Lahr, Gustavo Jose
and Caurin, Glauco A. P.",
editor="Pucheta, Mart{\'i}n
and Cardona, Alberto
and Preidikman, Sergio
and Hecker, Rogelio",
title="In-Hand Manipulation via Deep Reinforcement Learning for Industrial Robots",
booktitle="Multibody Mechatronic Systems",
year="2021",
publisher="Springer International Publishing",
address="Cham",
pages="222--228",
}

@article{kang2021high,
  title={High-Speed Autonomous Robotic Assembly Using In-Hand Manipulation and Re-Grasping},
  author={Kang, Taewoong and Yi, Jae-Bong and Song, Dongwoon and Yi, Seung-Joon},
  journal={Applied Sciences},
  volume={11},
  number={1},
  pages={37},
  year={2021},
  publisher={Multidisciplinary Digital Publishing Institute}
}

@article{wang2022tacto,
  title={Tacto: A fast, flexible, and open-source simulator for high-resolution vision-based tactile sensors},
  author={Wang, Shaoxiong and Lambeta, Mike and Chou, Po-Wei and Calandra, Roberto},
  journal={IEEE Rob. \& Aut. Let.},
  volume={7},
  pages={3930--3937},
  year={2022},
}

@inproceedings{chi2023diffusionpolicy,
	title={Diffusion Policy: Visuomotor Policy Learning via Action Diffusion},
	author={Chi, Cheng and Feng, Siyuan and Du, Yilun and Xu, Zhenjia and Cousineau, Eric and Burchfiel, Benjamin and Song, Shuran},
	booktitle={Robotics: Science and Systems (RSS)},
	year={2023}
}

@ARTICLE{Monastirsky2023,
  author={Monastirsky, Maxim and Azulay, Osher and Sintov, Avishai},
  journal={IEEE Robotics and Automation Letters}, 
  title={Learning to Throw With a Handful of Samples Using Decision Transformers}, 
  year={2023},
  volume={8},
  number={2},
  pages={576-583},
  doi={10.1109/LRA.2022.3229266}}

@article{Lehman2010,
  title={Dexterous miniature robot for advanced minimally invasive surgery},
  author={Amy C. Lehman and Nathan A. Wood and Shane Farritor and Matthew R. Goede and Dmitry Oleynikov},
  journal={Surgical Endoscopy},
  year={2010},
  volume={25},
  pages={119-123},
  url={https://api.semanticscholar.org/CorpusID:32895824}
}

@Article{Stefanelli2023,
AUTHOR = {Stefanelli, Enrica and Cordella, Francesca and Gentile, Cosimo and Zollo, Loredana},
TITLE = {Hand Prosthesis Sensorimotor Control Inspired by the Human Somatosensory System},
JOURNAL = {Robotics},
VOLUME = {12},
YEAR = {2023},
NUMBER = {5},
ARTICLE-NUMBER = {136},
}

@article{Starke2022,
title = {Semi-autonomous control of prosthetic hands based on multimodal sensing, human grasp demonstration and user intention},
journal = {Robotics and Autonomous Systems},
volume = {154},
pages = {104123},
year = {2022},
issn = {0921-8890},
author = {Julia Starke and Pascal Weiner and Markus Crell and Tamim Asfour},
}

@article{Cordella2016,
  title={Literature Review on Needs of Upper Limb Prosthesis Users},
  author={Francesca Cordella and Anna Lisa Ciancio and Rinaldo Sacchetti and Angelo Davalli and Andrea Giovanni Cutti and Eugenio Guglielmelli and Loredana Zollo},
  journal={Frontiers in Neuroscience},
  year={2016},
  volume={10},
}

@article{Marinelli2022,
author = {Marinelli, A. and Boccardo, Nicolò and Tessari, Federico and Di Domenico, Dario and Caserta, Giulia and Canepa, Michele and Gini, Giuseppina and Barresi, Giacinto and Laffranchi, Matteo and De Michieli, Lorenzo and Semprini, Marianna},
year = {2022},
month = {12},
pages = {},
title = {Active upper limb prostheses: A review on current state and upcoming breakthroughs},
volume = {5},
journal = {Progress in Biomedical Engineering},
}

@article{Petrich2021,
      title={Assistive arm and hand manipulation: How does current research intersect with actual healthcare needs?}, 
      author={Laura Petrich and Jun Jin and Masood Dehghan and Martin Jagersand},
      year={2021},
      archivePrefix={arXiv preprint arXiv:2101.02750},
      primaryClass={cs.RO}
}

@inproceedings{korthals2019multisensory,
  title={Multisensory assisted in-hand manipulation of objects with a dexterous hand},
  author={Korthals, Timo and Melnik, Andrew and Leitner, J{\"u}rgen and Hesse, Marc},
  booktitle={Proceedings of 2019 ICRA Workshop on ViTac: Integrating Vision and Touch for Multimodal and Cross-modal Perception},
  pages={1--2},
  year={2019}
}

@inproceedings{melnik2019tactile,
  title={Tactile sensing and deep reinforcement learning for in-hand manipulation tasks},
  author={Melnik, Andrew and Lach, Luca and Plappert, Matthias and Korthals, Timo and Haschke, Robert and Ritter, Helge},
  booktitle={IROS Workshop on Autonomous Object Manipulation},
  year={2019}
}

@article{melnik2021,
  title={Using tactile sensing to improve the sample efficiency and performance of deep deterministic policy gradients for simulated in-hand manipulation tasks},
  author={Melnik, Andrew and Lach, Luca and Plappert, Matthias and Korthals, Timo and Haschke, Robert and Ritter, Helge},
  journal={Frontiers in Robotics and AI},
  pages={57},
  year={2021},
  publisher={Frontiers}
}

@inproceedings{kalashnikov2018scalable,
  title={Scalable deep reinforcement learning for vision-based robotic manipulation},
  author={Kalashnikov, Dmitry and Irpan, Alex and Pastor, Peter and Ibarz, Julian and Herzog, Alexander and Jang, Eric and Quillen, Deirdre and Holly, Ethan and Kalakrishnan, Mrinal and Vanhoucke, Vincent and others},
  booktitle={Conference on Robot Learning},
  pages={651--673},
  year={2018},
  organization={PMLR}
}

@inproceedings{eysenbach2018leave,
  title={Leave no Trace: Learning to Reset for Safe and Autonomous Reinforcement Learning},
  author={Eysenbach, B and Gu, S and Ibarz, J and Levine, S},
  booktitle={International Conference on Learning Representations (ICLR)},
  year={2018},
  organization={OpenReview. net}
}

@article{srinivasan2020learning,
  title={Learning to be safe: Deep {RL} with a safety critic},
  author={Srinivasan, Krishnan and Eysenbach, Benjamin and Ha, Sehoon and Tan, Jie and Finn, Chelsea},
  journal={arXiv preprint arXiv:2010.14603},
  year={2020}
}

@inproceedings{zhao2020sim,
  title={Sim-to-real transfer in deep reinforcement learning for robotics: a survey},
  author={Zhao, Wenshuai and Queralta, Jorge Pe{\~n}a and Westerlund, Tomi},
  booktitle={IEEE Symposium Series on Computational Intelligence (SSCI)},
  pages={737--744},
  year={2020},
}

@inproceedings{van2019sim,
  title={Sim-to-real transfer learning using robustified controllers in robotic tasks involving complex dynamics},
  author={Van Baar, Jeroen and Sullivan, Alan and Cordorel, Radu and Jha, Devesh and Romeres, Diego and Nikovski, Daniel},
  booktitle={IEEE International Conference on Robotics and Automation (ICRA)},
  pages={6001--6007},
  year={2019}
}

@article{psomopoulou2021robust,
  title={A robust controller for stable 3d pinching using tactile sensing},
  author={Psomopoulou, Efi and Pestell, Nicholas and Papadopoulos, Fotios and Lloyd, John and Doulgeri, Zoe and Lepora, Nathan F},
  journal={IEEE Robotics and Automation Letters},
  volume={6},
  number={4},
  pages={8150--8157},
  year={2021},
  publisher={IEEE}
}

@inproceedings{morgan2021model,
  title={Model predictive actor-critic: accelerating robot skill acquisition with deep reinforcement learning},
  author={Morgan, Andrew S and Nandha, Daljeet and Chalvatzaki, Georgia and D’Eramo, Carlo and Dollar, Aaron M and Peters, Jan},
  booktitle={IEEE International Conference on Robotics and Automation (ICRA)},
  pages={6672--6678},
  year={2021},
}

@article{lambeta2020digit,
  title={DIGIT: A novel design for a low-cost compact high-resolution tactile sensor with application to in-hand manipulation},
  author={Lambeta, Mike and Chou, Po-Wei and Tian, Stephen and Yang, Brian and Maloon, Benjamin and Most, Victoria Rose and Stroud, Dave and Santos, Raymond and Byagowi, Ahmad and Kammerer, Gregg and others},
  journal={IEEE Robotics and Automation Letters},
  volume={5},
  number={3},
  pages={3838--3845},
  year={2020},
  publisher={IEEE}
}

@inproceedings{wang2020swingbot,
  title={Swingbot: Learning physical features from in-hand tactile exploration for dynamic swing-up manipulation},
  author={Wang, Chen and Wang, Shaoxiong and Romero, Branden and Veiga, Filipe and Adelson, Edward},
  booktitle={IEEE/RSJ International Conference on Intelligent Robots and Systems (IROS)},
  pages={5633--5640},
  year={2020}
}

@ARTICLE{Veiga2020,
AUTHOR={Veiga, Filipe and Akrour, Riad and Peters, Jan},   
TITLE={Hierarchical Tactile-Based Control Decomposition of Dexterous In-Hand Manipulation Tasks},  
JOURNAL={Frontiers in Robotics and AI},      
VOLUME={7},           
YEAR={2020},      
URL={https://www.frontiersin.org/articles/10.3389/frobt.2020.521448},       
DOI={10.3389/frobt.2020.521448},      
ISSN={2296-9144},   
}

@article{azulay2022learning,
  author={Azulay, Osher and Ben-David, Inbar and Sintov, Avishai},
  journal={IEEE Transactions on Haptics}, 
  title={Learning Haptic-Based Object Pose Estimation for In-Hand Manipulation Control With Underactuated Robotic Hands}, 
  year={2022},
  volume={},
  number={},
  pages={1-12},
  doi={10.1109/TOH.2022.3232713}}

@article{azulay2022haptic,
  title={Haptic-Based and $ SE (3) $-Aware Object Insertion Using Compliant Hands},
  author={Azulay, Osher and Monastirsky, Maxim and Sintov, Avishai},
  journal={IEEE Robotics and Automation Letters},
  volume={8},
  number={1},
  pages={208--215},
  year={2022},
  publisher={IEEE}
}

@inproceedings{kimmel2019belief,
  title={Belief-space planning using learned models with application to underactuated hands},
  author={Kimmel, Andrew and Sintov, Avishai and Tan, Juntao and Wen, Bowen and Boularias, Abdeslam and Bekris, Kostas E},
  booktitle={International Symposium on Robotics Research (ISRR)},
  year={2019}
}

@inproceedings{Ma2016,
    author = {Ma, Raymond R. and Dollar, Aaron M.},
    title = "{In-Hand Manipulation Primitives for a Minimal, Underactuated Gripper With Active Surfaces}",
    volume = {Volume 5A: 40th Mechanisms and Robotics Conference},
    booktitle = {ASME International Design Engineering Technical Conferences and Computers and Information in Engineering Conference},
    year = {2016},
    month = {08},
}

@article{Chen2021,
author={Chen, Yuan and Prepscius, Colin and Lee, Daewon and Lee, Daniel D.},
  journal={IEEE Robotics and Automation Letters}, 
  title={Tactile Velocity Estimation for Controlled In-Grasp Sliding}, 
  year={2021},
  volume={6},
  number={2},
  pages={1614-1621}}

@article{Yuan2023,
      title={Robot Synesthesia: In-Hand Manipulation with Visuotactile Sensing}, 
      author={Ying Yuan and Haichuan Che and Yuzhe Qin and Binghao Huang and Zhao-Heng Yin and Kang-Won Lee and Yi Wu and Soo-Chul Lim and Xiaolong Wang},
      year={2023},
      primaryClass={cs.RO},
    journal={arXiv preprint arXiv:2312.01853}
}

@ARTICLE{Bellman1957,
    author = "Richard Bellman",
     title = "A Markovian Decision Process",
   journal = "Ind. Uni. Math. J.",
  fjournal = "Indiana University Mathematics Journal",
    volume = 6,
      year = 1957,
     issue = 4,
     pages = "679--684",
      issn = "0022-2518",
     coden = "IUMJAB",
   mrclass = "",
}

@article{Fonseca2019,
author = {Fonseca, Vinicius and Alves de Oliveira, Thiago Eustaquio and Petriu, Emil},
year = {2019},
pages = {2285},
title = {Estimating the Orientation of Objects from Tactile Sensing Data Using Machine Learning Methods and Visual Frames of Reference},
journal = {Sensors}}

@INPROCEEDINGS{Zeng2018,
  author={Zeng, Andy and Song, Shuran and Yu, Kuan-Ting and Donlon, Elliott and Hogan, Francois R. and Bauza, Maria and Ma, Daolin and Taylor, Orion and Liu, Melody and Romo, Eudald and Fazeli, Nima and Alet, Ferran and Dafle, Nikhil Chavan and Holladay, Rachel and Morena, Isabella and Qu Nair, Prem and Green, Druck and Taylor, Ian and Liu, Weber and Funkhouser, Thomas and Rodriguez, Alberto},
  booktitle={IEEE International Conference on Robotics and Automation (ICRA)}, 
  title={Robotic Pick-and-Place of Novel Objects in Clutter with Multi-Affordance Grasping and Cross-Domain Image Matching}, 
  year={2018},
  volume={},
  number={},
  pages={3750-3757}}

@INPROCEEDINGS{Taylor2020,
  author={Taylor, Ian H. and Chavan-Dafle, Nikhil and Li, Godric and Doshi, Neel and Rodriguez, Alberto},
  booktitle={IEEE/RSJ International Conference on Intelligent Robots and Systems (IROS)}, 
  title={PnuGrip: An Active Two-Phase Gripper for Dexterous Manipulation}, 
  year={2020},
  volume={},
  number={},
  pages={9144-9150}}

@INPROCEEDINGS{Costanzo2021,
  author={Costanzo, Marco and De Maria, Giuseppe and Natale, Ciro},
  booktitle={International Conference on Advanced Robotics (ICAR)}, 
  title={Dual-Arm In-Hand Manipulation with Parallel Grippers Using Tactile Feedback}, 
  year={2021},
  volume={},
  number={},
  pages={942-947}}

@article{Costanzo2021b,
title = {Control of robotic object pivoting based on tactile sensing},
journal = {Mechatronics},
volume = {76},
pages = {102545},
year = {2021},
author = {Marco Costanzo},
}

@INPROCEEDINGS{Cruciani2018,
  author={Cruciani, Silvia and Smith, Christian},
  booktitle={IEEE/RSJ International Conference on Intelligent Robots and Systems (IROS)}, 
  title={Integrating Path Planning and Pivoting}, 
  year={2018},
  volume={},
  number={},
  pages={6601-6608}}

@INPROCEEDINGS{Cruciani2018b,
  author={Cruciani, Silvia and Smith, Christian and Kragic, Danica and Hang, Kaiyu},
  booktitle={IEEE/RSJ International Conference on Intelligent Robots and Systems (IROS)}, 
  title={Dexterous Manipulation Graphs}, 
  year={2018},
  volume={},
  number={},
  pages={2040-2047}}

@INPROCEEDINGS{Nagata1994,
  author={Nagata, Kazuyuki},
  booktitle={ IEEE International Conference on Robotics and Automation}, 
  title={Manipulation by a parallel-jaw gripper having a turntable at each fingertip}, 
  year={1994},
  volume={},
  number={},
  pages={1663-1670 vol.2}}

@INPROCEEDINGS{Zhao2020,
  title={Assembly of randomly placed parts realized by using only one robot arm with a general parallel-jaw gripper},
  author={Jie Zhao and Xin Jiang and Xiaoman Wang and Shengfan Wang and Yunhui Liu},
  journal={IEEE International Conference on Robotics and Automation (ICRA)},
  year={2020},
  pages={5024-5030}
}

@article{Zuo2021,
author = {Zuo, Shiping and Li, Jianfeng and Dong, Mingjie},
year = {2021},
month = {06},
pages = {1-16},
title = {Design, modeling, and manipulability evaluation of a novel four-DOF parallel gripper for dexterous in-hand manipulation},
volume = {35},
journal = {Journal of Mechanical Science and Technology},
}

@INPROCEEDINGS{Chapman2021,
  author={Chapman, Jayden and Gorjup, Gal and Dwivedi, Anany and Matsunaga, Saori and Mariyama, Toshisada and MacDonald, Bruce and Liarokapis, Minas},
  booktitle={IEEE International Conference on Robotics and Automation (ICRA)}, 
  title={A Locally-Adaptive, Parallel-Jaw Gripper with Clamping and Rolling Capable, Soft Fingertips for Fine Manipulation of Flexible Flat Cables}, 
  year={2021},
  volume={},
  number={},
  pages={6941-6947}}

@INPROCEEDINGS{Bimbo2013,
  author={Bimbo, Joao and Seneviratne, Lakmal D. and Althoefer, Kaspar and Liu, Hongbin},
  booktitle={IEEE/RSJ Int. Conf. on Intel. Rob. \& Sys}, 
  title={Combining touch and vision for the estimation of an object's pose during manipulation}, 
  year={2013},
  volume={},
  number={}}

@ARTICLE{Terasaki1998,
  author={Terasaki, Hajime and Hasegawa, Tsutomu},
  journal={IEEE Transactions on Robotics and Automation}, 
  title={Motion planning of intelligent manipulation by a parallel two-fingered gripper equipped with a simple rotating mechanism}, 
  year={1998},
  volume={14},
  number={2},
  pages={207-219}}

@article{Tegin2005,
author = {Tegin, Johan and Wikander, Jan},
year = {2005},
pages = {},
title = {Tactile sensing in intelligent robotic manipulation—A review},
volume = {32},
journal = {Indus. Robot: An Int. Jou.},
doi = {10.1108/01439910510573318}
}

@article{Lepora2020,
title = "Optimal Deep Learning for Robot Touch: Training Accurate Pose Models of 3D Surfaces and Edges",
author = "Nathan Lepora and John Lloyd",
year = "2020",
day = "7",
volume = "27",
journal = "IEEE Robotics and Automation Magazine",
number = "2",
}

@article{Su2012,
author = {Su, Zhe and Fishel, Jeremy and Yamamoto, Tomonori and Loeb, Gerald},
year = {2012},
month = {07},
pages = {7},
title = {Use of tactile feedback to control exploratory movements to characterize object compliance},
volume = {6},
journal = {Frontiers in neurorobotics},
}

@INPROCEEDINGS{Morgan2021,
      title={Vision-driven Compliant Manipulation for Reliable, High-Precision Assembly Tasks}, 
      author={Andrew S. Morgan and Bowen Wen and Junchi Liang and Abdeslam Boularias and Aaron M. Dollar and Kostas Bekris},
      year={2021},
      booktitle={Robotics: Science and Systems}
}

@inproceedings{sintov2020motion,
  title={Motion planning with competency-aware transition models for underactuated adaptive hands},
  author={Sintov, Avishai and Kimmel, Andrew and Bekris, Kostas E and Boularias, Abdeslam},
  booktitle={2020 IEEE International Conference on Robotics and Automation (ICRA)},
  pages={7761--7767},
  year={2020},
  organization={IEEE}
}

@Article{SintovInHand2017,
  Title                    = {Dynamic regrasping by in-hand orienting of grasped objects using non-dexterous robotic grippers},
  Author                   = {Avishai Sintov and Amir Shapiro},
  Journal                  = {Robotics and Computer-Integrated Manufacturing},
  Pages                    = {114 - 131},
  Volume                   = {50},
    year = {2017},
  ISSN                     = {0736-5845}
}

@Article{SintovSwingUp2016,
  Title                    = {Robotic Swing-up Regrasping Manipulation based on Impulse-Momentum approach and {cLQR} control},
  Author                   = {Avishai Sintov and Or Tslil and Amir Shapiro},
  Journal                  = {IEEE Transactions on Robotics},
  Number                   = {5},
  Pages                    = {1079-1090},
  Volume                   = {32},
year = {2016}
}

@article{morgan2020object,
  title={Object-agnostic dexterous manipulation of partially constrained trajectories},
  author={Morgan, Andrew S and Hang, Kaiyu and Dollar, Aaron M},
  journal={IEEE Robotics and Automation Letters},
  volume={5},
  number={4},
  pages={5494--5501},
  year={2020},
  publisher={IEEE}
}

@inproceedings{morgan2019data,
  title={A data-driven framework for learning dexterous manipulation of unknown objects},
  author={Morgan, Andrew S and Hang, Kaiyu and Bircher, Walter G and Dollar, Aaron M},
  booktitle={IEEE/RSJ International Conference on Intelligent Robots and Systems (IROS)},
  pages={8273--8280},
  year={2019},
}

@inproceedings{calli2018path,
  title={Path planning for within-hand manipulation over learned representations of safe states},
  author={Calli, Berk and Kimmel, Andrew and Hang, Kaiyu and Bekris, Kostas and Dollar, Aaron},
  booktitle={International Symposium on Experimental Robotics},
  pages={437--447},
  year={2018},
  organization={Springer}
}

@inproceedings{calli2016vision,
  title={Vision-based precision manipulation with underactuated hands: Simple and effective solutions for dexterity},
  author={Calli, Berk and Dollar, Aaron M},
  booktitle={IEEE/RSJ International Conference on Intelligent Robots and Systems (IROS)},
  pages={1012--1018},
  year={2016},
}

@article{deng2020learning,
  title={Learning synergies based in-hand manipulation with reward shaping},
  author={Deng, Zhen and Zhang, Jian Wei},
  journal={CAAI Transactions on Intelligence Technology},
  volume={5},
  number={3},
  pages={141--149},
  year={2020},
  publisher={Wiley Online Library}
}

@inproceedings{funabashi2020stable,
  title={Stable in-grasp manipulation with a low-cost robot hand by using 3-axis tactile sensors with a CNN},
  author={Funabashi, Satoshi and Isobe, Tomoki and Ogasa, Shun and Ogata, Tetsuya and Schmitz, Alexander and Tomo, Tito Pradhono and Sugano, Shigeki},
  booktitle={IEEE/RSJ International Conference on Intelligent Robots and Systems (IROS)},
  pages={9166--9173},
  year={2020},
}

@ARTICLE{Cruciani2020,
  author={Cruciani, Silvia and Sundaralingam, Balakumar and Hang, Kaiyu and Kumar, Vikash and Hermans, Tucker and Kragic, Danica},
  journal={IEEE Robotics and Automation Letters}, 
  title={Benchmarking In-Hand Manipulation}, 
  year={2020},
  volume={5},
  number={2},
  pages={588-595},
  doi={10.1109/LRA.2020.2964160}}

@INPROCEEDINGS{Sievers2022,
  author={Sievers, Leon and Pitz, Johannes and Bäuml, Berthold},
  booktitle={IEEE International Conference on Robotics and Automation (ICRA)}, 
  title={Learning Purely Tactile In-Hand Manipulation with a Torque-Controlled Hand}, 
  year={2022},
  volume={},
  number={},
  pages={2745-2751},
  doi={10.1109/ICRA46639.2022.9812093}}

@INPROCEEDINGS{Pitz2023,
  author={Pitz, Johannes and Röstel, Lennart and Sievers, Leon and Bäuml, Berthold},
  booktitle={IEEE International Conference on Robotics and Automation (ICRA)}, 
  title={Dextrous Tactile In-Hand Manipulation Using a Modular Reinforcement Learning Architecture}, 
  year={2023},
  volume={},
  number={},
  pages={1852-1858},
  doi={10.1109/ICRA48891.2023.10160756}}

@article{makoviychuk2021isaac,
      title={Isaac Gym: High Performance GPU-Based Physics Simulation For Robot Learning}, 
      author={Viktor Makoviychuk and Lukasz Wawrzyniak and Yunrong Guo and Michelle Lu and Kier Storey and Miles Macklin and David Hoeller and Nikita Rudin and Arthur Allshire and Ankur Handa and Gavriel State},
      year={2021},
      journal={arXiv preprint arXiv:2108.10470}
}

@article{Suomalainen2022,
title = {A survey of robot manipulation in contact},
journal = {Robotics and Autonomous Systems},
volume = {156},
pages = {104224},
year = {2022},
issn = {0921-8890},
doi = {https://doi.org/10.1016/j.robot.2022.104224},
url = {https://www.sciencedirect.com/science/article/pii/S0921889022001312},
author = {Markku Suomalainen and Yiannis Karayiannidis and Ville Kyrki},
}

@Article{Han2023,
AUTHOR = {Han, Dong and Mulyana, Beni and Stankovic, Vladimir and Cheng, Samuel},
TITLE = {A Survey on Deep Reinforcement Learning Algorithms for Robotic Manipulation},
JOURNAL = {Sensors},
VOLUME = {23},
YEAR = {2023},
NUMBER = {7},
ARTICLE-NUMBER = {3762},
}

@ARTICLE{Papadopoulos2021,
AUTHOR={Papadopoulos, Evangelos  and Aghili, Farhad  and Ma, Ou  and Lampariello, Roberto },
TITLE={Robotic Manipulation and Capture in Space: A Survey},
JOURNAL={Frontiers in Robotics and AI},
VOLUME={8},
YEAR={2021},
}

@INPROCEEDINGS{Herguedas2019,
  author={Herguedas, Rafael and López-Nicolás, Gonzalo and Aragüés, Rosario and Sagüés, Carlos},
  booktitle={IEEE International Conference on Emerging Technologies and Factory Automation (ETFA)}, 
  title={Survey on multi-robot manipulation of deformable objects}, 
  year={2019},
  volume={},
  number={},
  pages={977-984},
  doi={10.1109/ETFA.2019.8868987}}

@Article{Mohammed2022,
AUTHOR = {Mohammed, Marwan Qaid and Kwek, Lee Chung and Chua, Shing Chyi and Al-Dhaqm, Arafat and Nahavandi, Saeid and Eisa, Taiseer Abdalla Elfadil and Miskon, Muhammad Fahmi and Al-Mhiqani, Mohammed Nasser and Ali, Abdulalem and Abaker, Mohammed and Alandoli, Esmail Ali},
TITLE = {Review of Learning-Based Robotic Manipulation in Cluttered Environments},
JOURNAL = {Sensors},
VOLUME = {22},
YEAR = {2022},
NUMBER = {20},
ARTICLE-NUMBER = {7938},
URL = {https://www.mdpi.com/1424-8220/22/20/7938},
PubMedID = {36298284},
ISSN = {1424-8220},
DOI = {10.3390/s22207938}
}

@article{Billard2019,
author = {Aude Billard  and Danica Kragic },
title = {Trends and challenges in robot manipulation},
journal = {Science},
volume = {364},
number = {6446},
pages = {eaat8414},
year = {2019},
doi = {10.1126/science.aat8414},
URL = {https://www.science.org/doi/abs/10.1126/science.aat8414},
eprint = {https://www.science.org/doi/pdf/10.1126/science.aat8414},
}

@article{Cui2021,
author = {Jinda Cui  and Jeff Trinkle },
title = {Toward next-generation learned robot manipulation},
journal = {Science Robotics},
volume = {6},
number = {54},
pages = {eabd9461},
year = {2021},
}

@article{Feng2020,
title = {An overview of collaborative robotic manipulation in multi-robot systems},
journal = {Annual Reviews in Control},
volume = {49},
pages = {113-127},
year = {2020},
issn = {1367-5788},
doi = {https://doi.org/10.1016/j.arcontrol.2020.02.002},
url = {https://www.sciencedirect.com/science/article/pii/S1367578820300043},
author = {Zhi Feng and Guoqiang Hu and Yajuan Sun and Jeffrey Soon},
}

@article{Fang2019,
  title={Survey of imitation learning for robotic manipulation},
  author={Bin Fang and Shi-Dong Jia and Di Guo and Muhua Xu and Shuhuan Wen and Fuchun Sun},
  journal={International Journal of Intelligent Robotics and Applications},
  year={2019},
  volume={3},
  pages={362 - 369},
  url={https://api.semanticscholar.org/CorpusID:202733441}
}

@ARTICLE{Zhou2018,
  author={Zhou, Jianshu and Yi, Juan and Chen, Xiaojiao and Liu, Zixie and Wang, Zheng},
  journal={IEEE Robotics and Automation Letters}, 
  title={BCL-13: A 13-DOF Soft Robotic Hand for Dexterous Grasping and In-Hand Manipulation}, 
  year={2018},
  volume={3},
  number={4},
  pages={3379-3386},
  doi={10.1109/LRA.2018.2851360}}

@ARTICLE{Abondance2020,
  author={Abondance, Sylvain and Teeple, Clark B. and Wood, Robert J.},
  journal={IEEE Robotics and Automation Letters}, 
  title={A Dexterous Soft Robotic Hand for Delicate In-Hand Manipulation}, 
  year={2020},
  volume={5},
  number={4},
  pages={5502-5509},
  doi={10.1109/LRA.2020.3007411}}

@article{odhner2015stable,
  title={Stable, open-loop precision manipulation with underactuated hands},
  author={Odhner, Lael U and Dollar, Aaron M},
  journal={The International Journal of Robotics Research},
  volume={34},
  number={11},
  pages={1347--1360},
  year={2015},
  publisher={SAGE Publications Sage UK: London, England}
}

@article{hofer2021sim2real,
  title={Sim2Real in robotics and automation: Applications and challenges},
  author={H{\"o}fer, Sebastian and Bekris, Kostas and Handa, Ankur and Gamboa, Juan Camilo and Mozifian, Melissa and Golemo, Florian and Atkeson, Chris and Fox, Dieter and Goldberg, Ken and Leonard, John and others},
  journal={IEEE Transactions on automation, Science and engineering},
  volume={18},
  number={2},
  pages={398--400},
  year={2021},
}

@article{funabashi2019morphology,
  title={Morphology specific stepwise learning of in-hand manipulation with a four-fingered hand},
  author={Funabashi, Satoshi and Schmitz, Alexander and Ogasa, Shun and Sugano, Shigeki},
  journal={IEEE Transactions on Industrial Informatics},
  volume={16},
  number={1},
  pages={433--441},
  year={2019},
  publisher={IEEE}
}

@INPROCEEDINGS{Higo2018,
  author={Higo, Ryosuke and Yamakawa, Yuji and Senoo, Taku and Ishikawa, Masatoshi},
  booktitle={IEEE/RSJ International Conference on Intelligent Robots and Systems (IROS)}, 
  title={Rubik's Cube Handling Using a High-Speed Multi-Fingered Hand and a High-Speed Vision System}, 
  year={2018},
  volume={},
  number={},
  pages={6609-6614},
  doi={10.1109/IROS.2018.8593538}}

@inproceedings{funabashi2020variable,
  title={Variable in-hand manipulations for tactile-driven robot hand via CNN-LSTM},
  author={Funabashi, Satoshi and Ogasa, Shun and Isobe, Tomoki and Ogata, Tetsuya and Schmitz, Alexander and Tomo, Tito Pradhono and Sugano, Shigeki},
  booktitle={IEEE/RSJ International Conference on Intelligent Robots and Systems (IROS)},
  pages={9472--9479},
  year={2020}
}

@INPROCEEDINGS{Wang2018,
  author={Z. {Wang} and C. R. {Garrett} and L. P. {Kaelbling} and T. {Lozano-Pérez}},
  booktitle={IEEE/RSJ Int. Conf. on Int. Rob. \& Sys.}, 
  title={Active Model Learning and Diverse Action Sampling for Task and Motion Planning}, 
  year={2018},
  pages={4107-4114}}

@ARTICLE{Zeng2020,
  author={A. {Zeng} and S. {Song} and J. {Lee} and A. {Rodriguez} and T. {Funkhouser}},
  journal={IEEE Transactions on Robotics}, 
  title={TossingBot: Learning to Throw Arbitrary Objects With Residual Physics}, 
  year={2020},
  volume={36},
  number={4},
  pages={1307-1319},
  doi={10.1109/TRO.2020.2988642}}

@InProceedings{SintovL4DC2020,
  title = 	 {Tools for Data-driven Modeling of Within-Hand Manipulation with Underactuated Adaptive Hands},
  author = 	 {Avishai Sintov and Andrew Kimmel and Bowen Wen and Abdeslam Boularias and Kostas Bekris},
  booktitle = 	 {Proceedings of the Conf. on Learning for Dyn. \& Cont.},
  pages = 	 {771--780},
  year = 	 {2020},
  volume = 	 {120},
  Address                  = {Berkeley, CA, USA},
  year = {2020},
}

@inproceedings{zhu2019dexterous,
  title={Dexterous manipulation with deep reinforcement learning: Efficient, general, and low-cost},
  author={Zhu, Henry and Gupta, Abhishek and Rajeswaran, Aravind and Levine, Sergey and Kumar, Vikash},
  booktitle={IEEE International Conference on Robotics and Automation (ICRA)},
  pages={3651--3657},
  year={2019}
}

@inproceedings{allshire2022transferring,
  title={Transferring dexterous manipulation from {GPU} simulation to a remote real-world trifinger},
  author={Allshire, Arthur and Mittal, Mayank and Lodaya, Varun and Makoviychuk, Viktor and Makoviichuk, Denys and Widmaier, Felix and W{\"u}thrich, Manuel and Bauer, Stefan and Handa, Ankur and Garg, Animesh},
  booktitle={IEEE/RSJ international conference on intelligent robots and systems (IROS)},
  pages={11802--11809},
  year={2022}
}

@article{sun2021characterizing,
  title={Characterizing Continuous Manipulation Families for Dexterous Soft Robot Hands},
  author={Sun, Jiatian and King, Jonathan P and Pollard, Nancy S},
  journal={Frontiers in Robotics and AI},
  volume={8},
  pages={645290},
  year={2021},
  publisher={Frontiers Media SA}
}

@article{ichnowski2021dex,
  title={Dex-nerf: Using a neural radiance field to grasp transparent objects},
  author={Ichnowski, Jeffrey and Avigal, Yahav and Kerr, Justin and Goldberg, Ken},
  journal={arXiv preprint arXiv:2110.14217},
  year={2021}
}

@article{wang2021survey,
  title={A survey on curriculum learning},
  author={Wang, Xin and Chen, Yudong and Zhu, Wenwu},
  journal={IEEE Transactions on Pattern Analysis and Machine Intelligence},
  year={2021},
  publisher={IEEE}
}

@article{morrison2020egad,
  title={Egad! an evolved grasping analysis dataset for diversity and reproducibility in robotic manipulation},
  author={Morrison, Douglas and Corke, Peter and Leitner, J{\"u}rgen},
  journal={IEEE Robotics and Automation Letters},
  volume={5},
  number={3},
  pages={4368--4375},
  year={2020},
  publisher={IEEE}
}

@inproceedings{calli2015ycb,
  title={The ycb object and model set: Towards common benchmarks for manipulation research},
  author={Calli, Berk and Singh, Arjun and Walsman, Aaron and Srinivasa, Siddhartha and Abbeel, Pieter and Dollar, Aaron M},
  booktitle={International Conference on Advanced Robotics (ICAR)},
  pages={510--517},
  year={2015},
  organization={IEEE}
}

@article{yuan2017gelsight,
  title={Gelsight: High-resolution robot tactile sensors for estimating geometry and force},
  author={Yuan, Wenzhen and Dong, Siyuan and Adelson, Edward H},
  journal={Sensors},
  volume={17},
  number={12},
  pages={2762},
  year={2017},
  publisher={MDPI}
}

@inproceedings{zimmer2014teacher,
  title={Teacher-student framework: a reinforcement learning approach},
  author={Zimmer, Matthieu and Viappiani, Paolo and Weng, Paul},
  booktitle={AAMAS Workshop Autonomous Robots and Multirobot Systems},
  year={2014}
}

@INPROCEEDINGS{radosavovic2020state,
  author={Radosavovic, Ilija and Wang, Xiaolong and Pinto, Lerrel and Malik, Jitendra},
  booktitle={IEEE/RSJ International Conference on Intelligent Robots and Systems (IROS)}, 
  title={State-Only Imitation Learning for Dexterous Manipulation}, 
  year={2021},
  volume={},
  number={},
  pages={7865-7871}}

@INPROCEEDINGS{bhatt2022surprisingly, 
  author       = {Aditya Bhatt and
                  Adrian Sieler and
                  Steffen Puhlmann and
                  Oliver Brock},
  title        = {Surprisingly Robust In-Hand Manipulation: An Empirical Study},
  BOOKTITLE = {Robotics: Science and Systems}, 
  year         = {2021},
}

@INPROCEEDINGS{Arunachalam2023holo,
  title={Holo-Dex: Teaching Dexterity with Immersive Mixed Reality},
  author={Sridhar Pandian Arunachalam and Irmak Guzey and Soumith Chintala and Lerrel Pinto},
  booktitle={IEEE International Conference on Robotics and Automation (ICRA)},
  year={2023},
  pages={5962-5969}
}

@misc{cai2019global,
      title={On the Global Convergence of Imitation Learning: A Case for Linear Quadratic Regulator}, 
      author={Qi Cai and Mingyi Hong and Yongxin Chen and Zhaoran Wang},
      year={2019},
      eprint={1901.03674},
      archivePrefix={arXiv},
      primaryClass={cs.LG}
}

@INPROCEEDINGS{Orbik2021,
  author={Orbik, Jedrzej and Agostini, Alejandro and Lee, Dongheui},
  booktitle={IEEE International Conference on Development and Learning (ICDL)}, 
  title={Inverse reinforcement learning for dexterous hand manipulation}, 
  year={2021},
  volume={},
  number={},
  pages={1-7},
  doi={10.1109/ICDL49984.2021.9515637}}

@inproceedings{Arunachalam2023,
  author={Arunachalam, Sridhar Pandian and Silwal, Sneha and Evans, Ben and Pinto, Lerrel},
  booktitle={IEEE International Conference on Robotics and Automation (ICRA)}, 
  title={Dexterous Imitation Made Easy: A Learning-Based Framework for Efficient Dexterous Manipulation}, 
  year={2023},
  volume={},
  number={},
  pages={5954-5961},
  doi={10.1109/ICRA48891.2023.10160275}}

@InProceedings{Ross2011,
  title = 	 {A Reduction of Imitation Learning and Structured Prediction to No-Regret Online Learning},
  author = 	 {Ross, Stephane and Gordon, Geoffrey and Bagnell, Drew},
  booktitle = 	 {International Conference on Artificial Intelligence and Statistics},
  pages = 	 {627--635},
  year = 	 {2011},
  editor = 	 {Gordon, Geoffrey and Dunson, David and Dudík, Miroslav},
  volume = 	 {15},
  series = 	 {Proceedings of Machine Learning Research},
}

@article{Goodfellow2020,
author = {Goodfellow, Ian and Pouget-Abadie, Jean and Mirza, Mehdi and Xu, Bing and Warde-Farley, David and Ozair, Sherjil and Courville, Aaron and Bengio, Yoshua},
title = {Generative Adversarial Networks},
year = {2020},
issue_date = {November 2020},
publisher = {Association for Computing Machinery},
address = {New York, NY, USA},
volume = {63},
number = {11},
url = {https://doi.org/10.1145/3422622},
doi = {10.1145/3422622},
journal = {Commun. ACM},
pages = {139–144},
numpages = {6}
}

@article{Wei2023,
      title={In-Hand Re-grasp Manipulation with Passive Dynamic Actions via Imitation Learning}, 
      author={Dehao Wei and Guokang Sun and Zeyu Ren and Shuang Li and Zhufeng Shao and Xiang Li and Nikos Tsagarakis and Shaohua Ma},
      year={2023},
      journal={arXiv preprint arXiv:2309.15455},
      primaryClass={cs.RO}
}

@article{wei2023wearable,
      title={A Wearable Robotic Hand for Hand-over-Hand Imitation Learning}, 
      author={Dehao Wei and Huazhe Xu},
      year={2023},
        journal={arXiv preprint arXiv:2309.14860},
      primaryClass={cs.RO}
}

@article{Ho2016,
  author       = {Jonathan Ho and
                  Stefano Ermon},
  title        = {Generative Adversarial Imitation Learning},
  journal      = {CoRR},
  volume       = {abs/1606.03476},
  year         = {2016},
  url          = {http://arxiv.org/abs/1606.03476},
  eprinttype    = {arXiv},
  eprint       = {1606.03476},
  bibsource    = {dblp computer science bibliography, https://dblp.org}
}

@article{zheng2022imitation,
  title={Imitation learning: Progress, taxonomies and challenges},
  author={Zheng, Boyuan and Verma, Sunny and Zhou, Jianlong and Tsang, Ivor W and Chen, Fang},
  journal={IEEE Transactions on Neural Networks and Learning Systems},
  number={99},
  pages={1--16},
  year={2022},
  publisher={IEEE}
}

@INPROCEEDINGS{Hammond2012,
  author={Hammond, Frank L. and Weisz, Jonathan and de la Llera Kurth, Andrés A. and Allen, Peter K. and Howe, Robert D.},
  booktitle={2012 IEEE International Conference on Robotics and Automation}, 
  title={Towards a design optimization method for reducing the mechanical complexity of underactuated robotic hands}, 
  year={2012},
  volume={},
  number={},
  pages={2843-2850},
  doi={10.1109/ICRA.2012.6225010}}

@inproceedings{Molnar2022,
author = {Molnar, Jennifer and Menguc, Yigit},
title = {Toward Handling the Complexities of Non-Anthropomorphic Hands},
year = {2022},
isbn = {9781450391566},
publisher = {Association for Computing Machinery},
booktitle = {CHI Conference on Human Factors in Computing Systems},
articleno = {210},
numpages = {9},
location = {New Orleans, LA, USA},
}

\end{document}